\documentclass[10pt,twocolumn,letterpaper]{article}

\usepackage{0_config/cvpr}              %
\makeatletter
\@namedef{ver@everyshi.sty}{}
\makeatother
\usepackage{tikz}
\usepackage{graphicx}
\usepackage{amsmath}
\usepackage{amssymb}
\usepackage{booktabs}

\usepackage[pagebackref,breaklinks,colorlinks]{hyperref}

\usepackage{graphicx}
\usepackage{tikz}
\usepackage[accsupp]{axessibility}  %
\usepackage{enumitem} %
\usepackage{comment} %
\usepackage{multirow}
\usepackage{pifont}%
\usepackage{float}
\newcommand{\cmark}{\ding{51}}%
\newcommand{\xmark}{\ding{55}}%

\usepackage{enumitem}
\setitemize{noitemsep,topsep=0pt,parsep=0pt,partopsep=0pt}
\usepackage{balance}
\usepackage{dsfont}
\usepackage{lipsum}  
\usepackage[thinc]{esdiff}

\definecolor{mypurple}{RGB}{0, 128, 255}
\definecolor{dexcolor}{RGB}{0, 117, 4}
\definecolor{honotatecolor}{RGB}{1, 0, 253}

\excludecomment{comments}

\excludecomment{remove}

\newcommand{\blue}[1]{{\color{blue}#1}}
\newcommand{\red}[1]{{\color{red}#1}}

\newcommand{\colorRef}[1]{\textcolor{red}{#1}} %
\newcommand{\reffig}[1]{\colorRef{Fig.~\ref{#1}}}

\newcommand{\refFig}[1]{\mbox{\colorRef{Figure~\ref{#1}}}}

\newcommand{\refTab}[1]{\mbox{\colorRef{Table~\ref{#1}}}}

\newcommand{\refsec}[1]{\colorRef{Sec.~\ref{#1}}}

\newcommand{\ccite}[1]{~\cite{#1}}
\newcommand{\subtitle}[1]{\textbf{#1}.}

\definecolor{GreenColor}{rgb}{0.137,0.573,0.565}
\definecolor{OrangeColor}{rgb}{0.914,0.541,0.0.141}
\definecolor{PurpleColor}{rgb}{0.5,0,0.7}
\definecolor{BlueColor}{rgb}{0,0.725,0.949}
\definecolor{PinkColor}{rgb}{0.9843,0.19215,0.6}

\newcommand{\M}[1]{\mathbf{#1}} %
\newcommand{\V}[1]{\mathbf{#1}} %
\newcommand{\R}{\rm I\!R}

\newcommand{\norm}[1]{\left\lVert#1\right\rVert}

\DeclareMathOperator*{\argmin}{arg\,min}

\newcommand{\myparagraph}[1]{\noindent\textbf{#1:}}

\newcommand{\nameCOLOR}[1]{\textcolor{black}{#1}} %

\newcommand{\TITLE}{ARCTIC: A Dataset for Dexterous Bimanual Hand-Object Manipulation\vspace{-0.5 em}}

\newcommand{\datasetname}{\mbox{\nameCOLOR{ARCTIC}}\xspace}
\newcommand{\datasetfullname}{\textbf{A}\textbf{R}ticulated obje\textbf{C}\textbf{T}s in \textbf{I}ntera\textbf{C}tion}
\newcommand{\methodname}{\mbox{\nameCOLOR{ArcticNet}}\xspace}
\newcommand{\methodnameSF}{\mbox{\nameCOLOR{ArcticNet-SF}}\xspace}
\newcommand{\methodnameLSTM}{\mbox{\nameCOLOR{ArcticNet-LSTM}}\xspace}
\newcommand{\interfieldSF}{\mbox{\nameCOLOR{InterField-SF}}\xspace}
\newcommand{\interfieldLSTM}{\mbox{\nameCOLOR{InterField-LSTM}}\xspace}

\newcommand{\intermethod}{\mbox{\nameCOLOR{InterField}}\xspace}
\newcommand{\taskpose}{consistent motion reconstruction\xspace}
\newcommand{\taskfield}{interaction field estimation\xspace}
\newcommand{\taskPose}{Consistent motion reconstruction\xspace}
\newcommand{\taskField}{Interaction field estimation\xspace}

\newcommand{\highlightNUMB}[1]{\textcolor{black}{#1}} %

\newcommand{\numSeqs}{{\highlightNUMB{$339$}}\xspace}
\newcommand{\numObjs}{{\highlightNUMB{$11$}}\xspace}
\newcommand{\numSubs}{{\highlightNUMB{$10$}}\xspace}

\newcommand{\numAllo}{{\highlightNUMB{$8$}}\xspace}
\newcommand{\numEgo}{{\highlightNUMB{$1$}}\xspace}
\newcommand{\fps}{{\highlightNUMB{$30$}}\xspace}
\newcommand{\numImages}{{\highlightNUMB{$2.1$M}}\xspace}
\newcommand{\numVicon}{{\highlightNUMB{$54$}}\xspace}
\newcommand{\smallMarkersize}{{\highlightNUMB{$1.5$mm}}\xspace}
\newcommand{\mediumMarkersize}{{\highlightNUMB{$4.5$mm}}\xspace}

\newcommand{\suppl}{\textcolor{black}{SupMat}\xspace}

\newcommand{\rgb}{RGB\xspace}
\newcommand{\rgbD}{\mbox{RGB-D}\xspace}
\newcommand{\rgbd}{\rgbD}

\newcommand{\oneD}{{1D}\xspace}
\newcommand{\twoD}{{2D}\xspace}
\newcommand{\sixD}{{6D}\xspace}

\newcommand{\threeD}{\xspace{3D}\xspace}

\newcommand{\mocap}{\mbox{MoCap}\xspace}
\newcommand{\moshpp}{\mbox{MoSh++}\xspace}
\newcommand{\mosh}{\moshpp}

\newcommand{\tpose}{\mbox{T-Pose}\xspace}
\newcommand{\blender}{{Blender}\xspace}
\newcommand{\groundtruth}{{ground-truth}\xspace}
\newcommand{\vicon}{\mbox{Vicon}\xspace}

\newcommand{\smplx}{\mbox{SMPL-X}\xspace}
\newcommand{\smplX}{\smplx}

\newcommand{\mano}{\mbox{MANO}\xspace}
\newcommand{\grab}{\mbox{GRAB}\xspace}
\newcommand{\honotate}{\mbox{HO-3D}\xspace}

\newcommand{\tsne}{\mbox{T-SNE}\xspace}

\newcommand{\pointnet}{\mbox{PointNet}\xspace}

\usepackage[capitalize]{cleveref}
\crefname{section}{Sec.}{Secs.}
\Crefname{section}{Section}{Sections}
\Crefname{table}{Table}{Tables}
\crefname{table}{Tab.}{Tabs.}

\def\cvprPaperID{xx} %
\def\confName{CVPR}
\def\confYear{2023}

\begin{document}

\title{\TITLE} %

\author{
Zicong Fan$^{1,3}$ 
\quad Omid Taheri$^{3}$ 
\quad Dimitrios Tzionas$^{2}$ 
\quad Muhammed Kocabas$^{1,3}$ \\
\quad Manuel Kaufmann$^{1}$ 
\quad Michael J. Black$^{3}$ 
\quad Otmar Hilliges$^{1}$
 \\
 {
 \small
 $^1$ETH Z{\"u}rich, Switzerland \quad
 $^2$University of Amsterdam \quad
 $^3$Max Planck Institute for Intelligent Systems, T{\"u}bingen, Germany
 }
}

\twocolumn[{%
\renewcommand\twocolumn[1][]{#1}%
\maketitle
\begin{center}
  \newcommand{\teaserwidth}{\textwidth}
  \centerline{\includegraphics[width=1.0\linewidth]{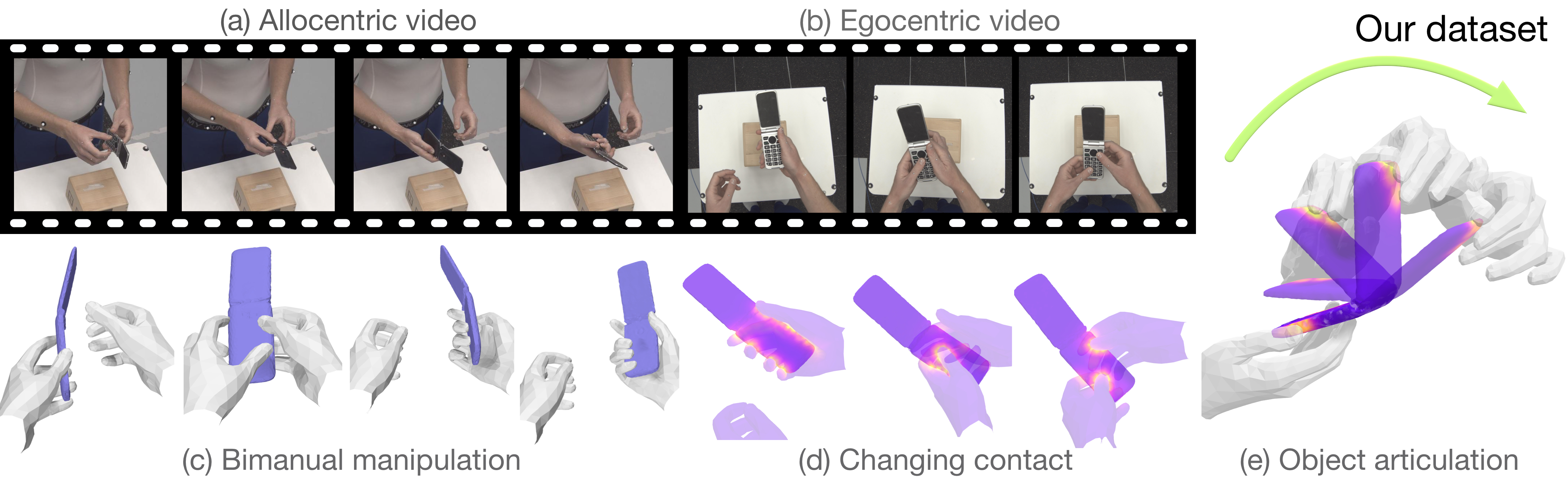}}
    \captionof{figure}{
    \datasetname is a dataset of hands dexterously manipulating {\em articulated} objects. 
    The dataset contains videos from both eight $3^{rd}$-person allocentric views (a) and one $1^{st}$-person egocentric view (b), together with accurate ground-truth 3D hand and object meshes, captured with a high-quality motion capture system.
    \datasetname goes beyond existing datasets to enable the study of dexterous bimanual manipulation of articulated objects (c) and provides detailed contact information between the hands and objects during manipulation (d-e).
    }
    \label{fig:teaser}
\end{center}%
}]

\maketitle

\begin{abstract}
\vspace{-4mm}
Humans intuitively understand that inanimate objects do not move by themselves, but that state changes are typically caused by human manipulation (\eg, the opening of a book). 
This is not yet the case for machines. 
In part this is because there exist no datasets with ground-truth 3D annotations for the study of physically consistent and synchronised motion of hands and articulated objects.
To this end,
we introduce \datasetname~-- a dataset of two hands that dexterously manipulate objects, containing \numImages video frames paired with accurate \threeD hand and object meshes and detailed, dynamic contact information.
It contains bi-manual articulation of objects such as scissors or laptops, where hand poses and object states evolve jointly in time.
We propose two novel articulated hand-object interaction tasks:
(1) \textit{\taskPose}: Given a monocular video, the goal is to reconstruct two hands and articulated objects in \threeD, so that their motions are spatio-temporally consistent.
(2) \textit{\taskField}: Dense relative hand-object distances must be estimated from images.
We introduce two baselines \methodname and \intermethod, respectively and evaluate them qualitatively and quantitatively on \datasetname.
Our code and data are available at \url{https://arctic.is.tue.mpg.de}.

\end{abstract}

\vspace{-8mm}
\section{Introduction}
\label{sec:intro}
\vspace{-2mm}
Humans constantly manipulate complex objects:
we open our laptop's cover to work, 
we apply spray to clean,
we carefully control our fingers to cut with scissors
-- rigid and articulated parts of objects move \emph{together} with our hands.
Inanimate objects only move or deform if external forces are applied to them.
The study of the physically consistent dynamics of hands and objects during manipulation has so far been under-researched in the hand pose estimation literature.
This is partly because existing hand-object datasets~\cite{hampali2020honnotate,hasson2019obman,dexycb,kwon2021h2o,hampali2022keypoint,liu2022hoi4d} are mostly limited to grasping of rigid objects and contain few if any examples of rich and dexterous manipulation of articulated objects.

To enable the study of dexterous articulated hand-object manipulation,
we collect a novel dataset called \datasetname (\datasetfullname).
\datasetname consists of video sequences of multi-view \rgb frames, and each frame is paired with accurate \threeD hand and object meshes. 
\datasetname contains data from \numSubs subjects interacting with \numObjs articulated objects, resulting in a total of \numImages \rgb images. 
Images are captured from multiple synchronized and calibrated views, including \numAllo static allocentric views and \numEgo moving egocentric view. 
To capture accurate \threeD meshes during manipulation, we %
synchronize color cameras with \numVicon high-resolution \vicon \mocap cameras\ccite{viconVantageWEB}.
These allow the use of small \mocap markers that do not interfere with hand-object interaction and are barely visible 
in the images.
We then fit pre-scanned human and object meshes to the observed markers\ccite{AMASS_2019,grab}. 
The objects consist of two rigid parts that rotate about a shared axis such as the flip phone in \reffig{fig:teaser} (for all objects, see \suppl).

Our dataset enables two novel tasks: 
(1) \taskpose,
(2) \taskfield.
For {\em \taskpose}, given 
a monocular video,
the task is to reconstruct the \threeD motion of two hands and an articulated object.
In particular, the reconstructed hand-object meshes should have spatio-temporally consistent hand-object contact, object articulation, and smooth motion during interaction. 
This task has several challenges:
\mbox{\highlightNUMB{(1)}} 
Spatio-temporal consistency requires precise hand-object \threeD alignment for all frames;
\mbox{\highlightNUMB{(2)}} 
This precision is hard to achieve due to depth ambiguity and severe occlusions during dexterous manipulation;
\mbox{\highlightNUMB{(3)}}
The unconstrained interaction causes more variations in hand pose and contact than in existing datasets~\cite{dexycb,hampali2020honnotate,hampali2022keypoint,liu2022hoi4d} (see \reffig{fig:tsne_contact}).

As an initial step towards addressing these challenges, and to provide baselines for future work, we introduce \methodname to reconstruct the motions of two hands and an articulated object from a video.
\methodname uses an encoder-decoder architecture to estimate parameters of the \mano hand model~\cite{mano} for the two hands, and our articulated object model.
We experiment with two variations of \methodname: a single-frame model and a temporal model with a recurrent architecture inspired by\ccite{kocabas2020vibe}.
We provide qualitative and quantitative results for future comparison.

When studying hand-object interaction, contact is important~\cite{grady2021contactopt,Yang_2021_CPF}. 
Some approaches~\cite{huang2022rich,Yang_2021_CPF} explore the task of binary contact estimation from a single \rgb image.
In the two-handed manipulation setting, hands can be near the object but not in contact.
To understand the dynamic, relative spatial configuration between hands and objects in more detail, even when not in contact, we propose the general task of \emph{\taskfield} from \rgb images. 
The goal is to estimate, for each hand vertex, the shortest distance to the object mesh and vice versa (see \reffig{fig:interfield_est} for a visualization).
We introduce a baseline, \intermethod, for this task and benchmark both a single-frame and a recurrent version of \intermethod on \datasetname for future comparison.

In summary, our contributions are as follows:
\highlightNUMB{(1)} 
We present \datasetname, the first large-scale dataset of two hands that \emph{dexterously} manipulate \emph{articulated} objects, with multi-view \rgb images paired with accurate \threeD meshes;
\highlightNUMB{(2)} 
We introduce two novel tasks of \taskpose and \taskfield to study the physically consistent motion of hands and articulated objects;
\highlightNUMB{(3)} 
We provide baselines for both tasks on \datasetname.

\section{Related Work}   
\label{sec:related}
\begin{table*}[t]
\centering
\vspace{-0.1in}
\resizebox{1.00\linewidth}{!}{
\begin{tabular}{lcccccccccccc}
\toprule
\multicolumn{1}{c}{dataset} & real   & \multicolumn{2}{c}{\# number of:}                    & ego-   & image      & articulated & both & human  & dexterous & annot.       \\ \cline{3-4}
 &
  \multicolumn{1}{c}{images} &
  \multicolumn{1}{c}{img} &
  \multicolumn{1}{c}{view} &
  \multicolumn{1}{c}{centric} &
  \multicolumn{1}{c}{resol.} &
  \multicolumn{1}{c}{objects} &
  \multicolumn{1}{c}{hands} &
  \multicolumn{1}{c}{body} &
  \multicolumn{1}{c}{manipulation} &
  \multicolumn{1}{c}{type} \\ \hline
FreiHand\ccite{Freihand2019}                    & \cmark & 37k  & 8                         & \xmark & 224$\times$224   & \xmark      & \xmark & \xmark & \xmark   & semi-auto    \\
ObMan\ccite{hasson2019obman}                       & \xmark & 154k & 1                       & \xmark & 256$\times$256   & \xmark      & \xmark& \xmark & \xmark   & synthetic    \\
FHPA\ccite{FirstPersonAction_CVPR2018}                        & \cmark & 105k & 1                        & \cmark & 1920$\times$1080 & \xmark      & \xmark & \xmark& \xmark   & magnetic     \\
HO3D\ccite{hampali2020honnotate}                        & \cmark & 78k  & 1-5                      & \xmark & 640$\times$480   & \xmark      & \xmark & \xmark& \xmark   & multi-kinect \\
ContactPose\ccite{contactpose_2020}                 & \cmark & 2.9M & 3                        & \xmark & 960$\times$540   & \xmark      & \xmark & \xmark& \xmark   & multi-kinect \\
GRAB\ccite{grab}                        & -      & -    & -                        & -      & -         & \xmark      & \cmark& \cmark & \xmark   & mocap        \\
DexYCB\ccite{dexycb}                      & \cmark & 582k & 8                        & \xmark & 640$\times$480   & \xmark      & \xmark& \xmark & \xmark   & multi-manual       \\
H2O\ccite{kwon2021h2o}                         & \cmark & 571k & 5                        & \cmark & 1280$\times$720  & \xmark      & \cmark& \xmark & \xmark   & multi-kinect \\ 
H2O-3D\ccite{hampali2022keypoint}                         & \cmark & 76k & 5                      & \xmark & 640$\times$480  & \xmark      & \cmark& \xmark & \xmark   & multi-kinect \\
HOI4D\ccite{liu2022hoi4d}                         & \cmark & 2.4M & 1                       & \cmark & 1280$\times$800  & \cmark      & \xmark& \xmark & \xmark   & single-manual \\ 
\textbf{\datasetname (Ours)}               & \cmark & \numImages & 9                        & \cmark & 2800$\times$2000 & \cmark      & \cmark & \cmark & \cmark  & mocap       \\ \bottomrule
\end{tabular}
}
\caption{\subtitle{Comparison of our \datasetname dataset with existing datasets} The keyword ``single/multi-manual" denotes whether single or multiple views being used to annotate manually.}
\label{tab:compare_datasets}
\end{table*}

\myparagraph{Human-object datasets}
Several datasets~\cite{mueller2017realtime,ballan2012motion,cao2021handobject,RealtimeHO_ECCV2016,tzionas2016articulated,tzionas2016capturing} contain images of human-object interaction, but here we focus on large-scale data~\cite{FirstPersonAction_CVPR2018,hampali2020honnotate,hasson2019obman,Freihand2019,sener2022assembly101,bhatnagar2022behave,huang2022intercap} that facilitates machine learning. 
There are three categories.
\textit{(1) Human body with rigid objects:}
Bhatnagar  \etal~\cite{bhatnagar2022behave} and Huang \etal~\cite{huang2022intercap} introduce image datasets for human body interaction with big objects.
Compared to ours,~\cite{bhatnagar2022behave} do not capture the hands.
Huang \etal~\cite{huang2022intercap} capture hands and body using a multi-view \rgbD setup while ours is captured using a \mocap setup for more accurate \threeD data.
Compared to both, we have dexterous bimanual manipulation, dynamic hand-object contact, and articulated objects.
GRAB~\cite{grab} contains detailed human-object interaction but no images, while BEDLAM~\cite{BEDLAM:CVPR:2023} contains videos with \groundtruth humans but no object interaction.
\textit{(2) Single hand with rigid objects:}
Most hand-object datasets~\cite{FirstPersonAction_CVPR2018,hampali2020honnotate,hasson2019obman,dexycb,liu2022hoi4d,contactpose_2020} consist of single-hand grasping interaction. However, hand poses in grasping interaction are mostly static, with relatively little pose variation over time.
Hampali \etal\ccite{hampali2020honnotate} use a multi-\rgbD system and fit both \mano and YCB object meshes with sequence-level fitting and contact constraints. 
\textit{(3) Two hands with rigid objects:}
Kwon \etal~\cite{kwon2021h2o} and Hampali \etal~\cite{hampali2022keypoint} present two-hand datasets interacting with rigid objects.
Compared to (2) and (3), our dataset has \threeD annotations of the full human body, both hands, and articulated objects.
We go beyond grasping and focus on less constrained dexterous bimanual manipulation.
We discuss the comparison between ours (\datasetname) and existing hand-object datasets~\cite{hampali2020honnotate,dexycb,kwon2021h2o,hampali2022keypoint,liu2022hoi4d} in \refsec{sec:compare_datasets}.

\myparagraph{Estimating \threeD hands and objects from \rgb images}
Monocular RGB 3D hand reconstruction has a long history since Rehg and Kanade~\cite{Rehg:1994}.
Most work in the literature focuses on hand-only reconstructions~\cite{iqbal2018hand,Mueller2018ganerated,Spurr2018crossmodal,spurr2021peclr,spurr2020eccv,zimmermann2017iccv,Boukhayma2019,hasson2019obman,ziani2022tempclr,ziani2022tempclr,fan2021digit,simon2017hand,Zhang2019endtoend,interhand,li2022interacting,zhang2021interacting,tzionas2013directional}.
Zimmermann \etal~\cite{zimmermann2017iccv} use a deep convolutional network for \threeD hand pose estimation via a multi-stage approach.
Spurr \etal~\cite{spurr2020eccv} introduce biomechanical constraints to regularize hand pose prediction.
Ziani \etal\ccite{ziani2022tempclr} use a self-supervised time-contrastive formulation to improve smoothness for hand motion reconstruction.
Recently, there has been increased interest in hand-object reconstruction from \rgb images~\cite{hasson2019obman,liu2021semi,Yang_2021_CPF,grady2021contactopt,Hasson2020photometric,tekin2019ho,corona2020ganhand,zhou2020monocular}.
Tekin \etal~\cite{tekin2019ho} infer \threeD control points for both the hand and the object in videos, using a temporal model to propagate information across time. 
Hasson \etal~\cite{hasson2019obman} render synthetic images and train a neural network to regress a static grasp of a \threeD hand and a rigid object, using full supervision together with contact losses. 
Corona 	\etal~\cite{corona2020ganhand} estimate \mano grasps for objects from an image, by first inferring the object shape and a rough hand pose, which is refined via contact constraints and an adversarial prior. 
Liu \etal~\cite{liu2021semi} use a transformer-based contextual-reasoning module that encodes the synergy between hand and object features, and has higher responses at contact regions.
Zhou \etal\ccite{zhou2022toch} learn an interaction motion prior to denoise motion predicted from an off-the-shelf single-frame hand-object reconstruction method. 
None of these methods deal with articulated objects, which result in complex hand-object interactions.

\myparagraph{Human-object contact detection}
Contact has been shown important for: 
pose taxonomies \cite{dillmann2005grasp,Feix_GRASP_2016,kamakura1980grasp}, 
pose estimation
\cite{grady2021contactopt,hampali2020honnotate,RealtimeHO_ECCV2016,Tsoli2018deformable,tzionas2016articulated,Yang_2021_CPF,hasson2019obman}, 
in-hand scanning
\cite{Tzionas2015inhand,zhang2021inhand}, and 
grasp synthesis \cite{grady2021contactopt,karunratanakul2020graspField,grab,Yang_2021_CPF}. %
Many methods \cite{grady2021contactopt,hampali2020honnotate,RealtimeHO_ECCV2016,Tsoli2018deformable,tzionas2016articulated} use the proximity between the \threeD hand/body and object meshes to estimate contacts and regularize pose estimation based on these.
Three main categories for contact estimation exist:
1) directly from meshes;
2) on the image pixel space from \rgb images;
3) binary contact in \threeD space from \rgb images.
Grady \etal~\cite{grady2021contactopt} take off-the-shelf regressors to estimate
grasping hand and object meshes, use these meshes to predict contacts on the objects provided by~\cite{contactpose_2020}, and leverage contacts to refine the grasp.
Their recent dataset~\cite{grady2022pressurevision} contains both contact and pressure between a hand and a flat sensor surface.
Tripathi \etal~\cite{tripathi2023ipman} infer pressure from body-scene contact.
Narasimhaswamy \etal~\cite{narasimhaswamy2020detecting} and Shan \etal~\cite{shan2020contact} infer bounding boxes for hands in contact on the input \rgb image. 
Chen \etal~\cite{chen2023hot} infer human-scene contact on pixels.
Rogez \etal~\cite{Rogez2015everyday} learn to infer contacts from the image using synthetic data, while Pham \etal~\cite{pham2018pami} use real contact data captured with instrumented objects.
Unlike others, \ccite{Rogez2015everyday} and~\cite{pham2018pami} estimate \threeD binary contact from \rgb images but the former does not generalize well to real images and the latter uses a classical approach due to the limited amount of data.
BSTRO estimates contact on the 3D body from an image but does not estimate 3D hand or object pose \cite{huang2022rich}.
Hi4D~\cite{yin2023hi4d} provides \groundtruth contact for close human interaction.
In contrast, our task of interaction field estimation goes beyond binary contact to model the dense relative distances between hands and objects.
Thanks to our dexterous manipulation,
\datasetname contains %
fast changing hand-object contact.

\section{\datasetname Dataset}

\myparagraph{Overview}
To allow the study of object articulation with hands in motion,
we construct \datasetname, a video dataset with accurate \threeD annotation for hands and articulated objects.
\datasetname contains \numSeqs sequences of dexterous manipulation of \numObjs articulated objects by \numSubs subjects 
(5 fe-/males).
The dataset consists of \numImages \rgb images from \numAllo static views and \numEgo egocentric view, paired with \threeD hand and object meshes.
To capture different interaction modes, we ask our subjects to either ``use" (1.7M images) or ``grasp" (457K images) the objects.
Depth images of the two hands, the human body, and objects can be rendered from \datasetname (see \suppl).

\subsection{Data Characteristics}
\label{sec:compare_datasets}

\myparagraph{Dataset features comparison} 
\refTab{tab:compare_datasets} compares \datasetname with existing hand-object datasets.
\datasetname is the only dataset that contains both hands, the full human body (in \smplx~\cite{pavlakos2019expressive}) and articulated objects.
\datasetname provides calibrated cameras ($8$ allocentric and $1$ egocentric) with high-resolution images, enabling the study of monocular, multi-view and egocentric reconstruction settings.
Importantly, \datasetname is a motion dataset that focuses on bimanual dexterous manipulation, meaning that subjects can freely interact with objects using both hands.
In contrast, existing hand-object datasets focus single-hand grasping~\cite{hampali2020honnotate,dexycb,hasson2019obman} and the movement is often controlled~\cite{hampali2022keypoint,kwon2021h2o}.
\grab~\cite{grab} has fast motion by using a similar \mocap setup but captures only rigid objects and does not have images.
HOI4D~\cite{liu2022hoi4d} is the only hand-object dataset that contains articulated objects, but it contains only a single view, does not capture the full human body, has a single hand, and mainly focuses on grasping.
Crucially, their hand data is captured from only a single egocentric view, 
which introduces ambiguity for the occluded fingers.

\myparagraph{Capture setup comparison}
Capturing dexterous manipulation while maintaining the quality of \threeD annotation is extremely challenging due to fast motion and heavy occlusion during the interaction.
In particular, the joints of a hand often have significant self-occlusion. 
The occlusion is even more severe when a hand interacts with objects and when there are multiple hands~\cite{interhand}.
Existing hand-object datasets\ccite{dexycb,hampali2020honnotate,liu2022hoi4d,kwon2021h2o,hampali2022keypoint} are captured with $1-8$ commodity \rgbd cameras, which is insufficient to eliminate occlusion.
As a result, their hand-object motion is often slow and they mainly focus on grasping interaction.
To reduce occlusion and to enable the capture of dexterous manipulation, we construct our dataset using an accurate \vicon \mocap setup with \numVicon high-end infrared Vantage-16 cameras\ccite{viconVantageWEB}.
To show our dexterous motion, and to compare \threeD annotation quality between datasets, see our project page video.

\begin{figure}[]
    \vspace{-4mm}
    \centering
    \includegraphics[width=1.0\linewidth]{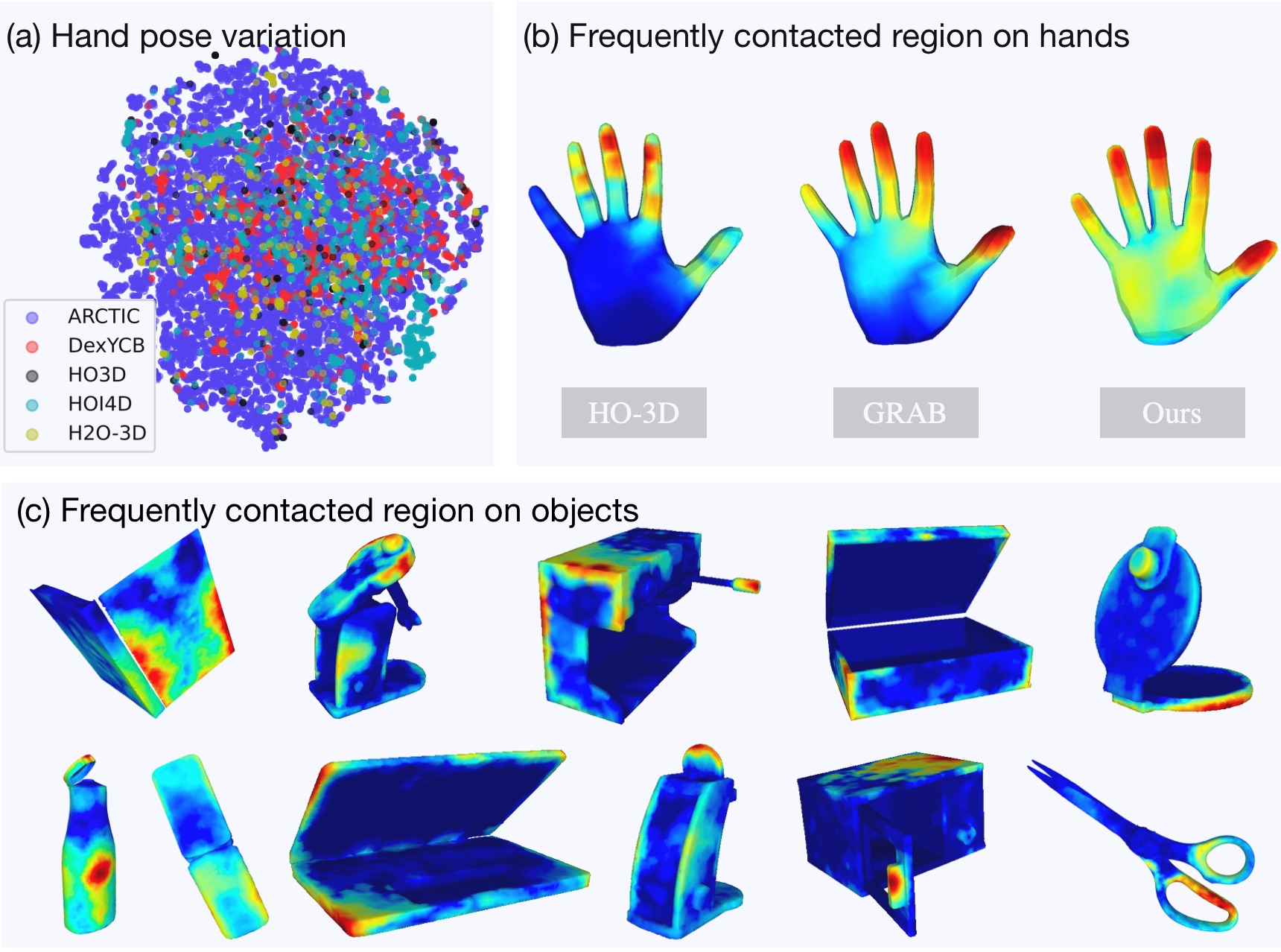}
    \vspace{-4mm}
    \caption{
    \subtitle{Hand pose and contact variations in datasets}
        (a) \tsne clustering of hand poses in different datasets.
        The plot shows that \datasetname has a significantly larger range of poses than all existing datasets.
        (b)
        Frequently contacted regions for hands in \honotate~\cite{hampali2020honnotate}, \grab~\cite{grab}, and \datasetname. 
        As seen with the broader heatmap spread on the hands, \datasetname has higher contact diversity.
        (c)
        Frequently contacted areas on our objects.
    }
    \vspace{-4mm}
    \label{fig:tsne_contact}
\end{figure}
\myparagraph{Hand pose and contact variations}
\refFig{fig:tsne_contact}\colorRef{a} compares different hand-object datasets~\cite{dexycb,hampali2020honnotate,liu2022hoi4d,hampali2022keypoint} in terms of hand pose variations by showing a \tsne clustering \cite{JMLR:v9:vandermaaten08a} of \threeD hand joints.
The plot reveals that our dataset (shown in \textcolor{blue}{blue}) has a significantly larger hand pose diversity than others. 
This is due to the unconstrained nature of \datasetname in which the subjects dexterously and dynamically {\em manipulate} the object (see project page video). 
The figure also shows frequently in-contact regions on hands (b) and objects (c) in the \datasetname dataset.
We generate the contact heatmaps following \grab's~\cite{grab} approach, by integrating per-frame binary contact labels for vertices over all sequences.
``Hotter'' regions denote a higher chance of being in contact while ``cooler'' regions denote lower chance of contact.
Similar to \honotate~\cite{hampali2020honnotate} and \grab~\cite{grab}, finger tips in our dataset are most likely to be in contact with objects.
However, thanks %
to the dexterous manipulation it contains, \datasetname has higher contact likelihood in the palm region than other datasets, hence the heatmaps appear more ``spread out''.
For regular-sized everyday objects, such as the ketchup bottle, the contact regions 
``agree'' with our usual interaction with them.
For smaller toy objects like the waffle iron, subjects are likely to pick up the object and support it with one hand, leading to ``hot'' regions on the bottom of the object.

\subsection{Acquisition Setup}

\label{sec:acquisition}

\begin{figure}[t]
    \centering
    \includegraphics[width=1.0\linewidth]{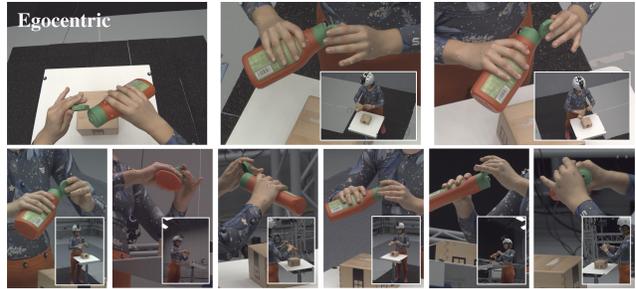}
    \caption{
    \subtitle{Our camera views} 
    We capture high resolution images in \numAllo static allocentric and \numEgo moving egocentric views. 
    Here we show zoomed-in crops and the original images.
    }
    \label{fig:camera_views}
\end{figure}

We detail our motion capture (\mocap) setup to acquire \threeD surfaces of strongly interacting hands and articulated objects.
We synchronize a \mocap system with a multi-view \rgb system. 
See \suppl for the marker sets.
With the latter we capture \rgb videos from \numAllo static allocentric views and \numEgo moving egocentric view at \fps FPS (see \reffig{fig:camera_views}). 
The capture pipeline has five steps: 
\highlightNUMB{(1)} obtaining the \threeD template geometry of the subjects and objects, 
\highlightNUMB{(2)} estimating the rotation axis for articulated objects, shown in  \suppl, 
\highlightNUMB{(3)} capturing interaction using marker-based \mocap together with calibrated and synchronized video, 
\highlightNUMB{(4)} solving for the poses of the body, hands, and objects from \mocap markers following~\cite{AMASS_2019,grab}, and 
\highlightNUMB{(5)} computing hand-object contact based on proximity, shown in \suppl.

\myparagraph{Obtaining canonical geometry}
We obtain the \groundtruth (GT) hand and body shape of each subject in a canonical \tpose using \threeD scans from a 3dMD\ccite{3dmdhand} scanner.
We register \smplX\ccite{pavlakos2019expressive} to \threeD scans at different time steps in varying poses 
and construct a personalized \threeD template of each subject.
See the \suppl for details of the template creation.
To obtain object geometries, we scan each object using an Artec \threeD hand-held scanner in a pre-defined pose.
We separate each scanned object mesh into two articulated parts in \blender.
See \suppl for all \numObjs articulated objects.

\myparagraph{Capturing human-object interaction}
To ensure accuracy, we perform full-body, hand and object tracking using a \vicon \mocap system with \numVicon infrared \mbox{Vantage-16} cameras\ccite{viconVantageWEB} to minimize the issues with occlusion.
To capture usable \rgb images alongside the \mocap data, we balance the trade-off between accuracy and marker intrusiveness by using small hemispherical markers with \smallMarkersize radius on the hands and objects.
The markers are placed on the dorsal side of the hand to not encumber participants during natural hand-object interaction, similar to \grab\ccite{grab}.
While our focus is on hands, we retrieve full-body pose estimates as they provide more reliable 
global rotations and translations for each hand. Therefore, we fit \smplx\ccite{pavlakos2019expressive} to the observed markers to attain realistic wrist articulations, as MANO contains no wrist articulation.

\myparagraph{Obtaining surfaces from \mocap}
Following \ccite{AMASS_2019,grab}, we associate \mocap marker positions with their corresponding subject/object vertices in the geometries obtained in canonical spaces.
We first pick initial guesses of marker-to-vertex correspondence on the subject/object meshes and use \moshpp~\cite{AMASS_2019} to refine the correspondence.
To obtain the full-body and hand surface that explain the \mocap data, we optimize \smplx pose using each subject's \smplX template to minimize the distance between the markers and their correspondences on the \smplx mesh.

The articulated object surface is parameterized by the \sixD pose of each object's base part and an 
\oneD
articulation relative to a canonical pose.
We obtain the \sixD pose of the object base for each \mocap frame by solving for the rigid transformation between the \mocap markers of the object base at that frame, and the object vertices corresponding to the markers in the object canonical space.
The 
\oneD
articulation is computed according to the estimated rotation axis (see \suppl) and a pre-defined rest pose.

\section{Evaluation Protocol}
\label{sec:eval_protocol}

\myparagraph{Data split}
We split the data by subjects, $8$ subjects for training, $1$ for validation (male) and $1$ for testing (female).
To ensure gender balance in evaluation, we use one male and one female subject.
With this same split, we establish two protocols: an allocentric protocol (\textbf{allo}) and an egocentric protocol (\textbf{ego}).
The former protocol lets us study our tasks in the  3rd-person, while the latter is similar to 1st-person views in a mixed-reality setting.
In the allocentric protocol, during training and evaluation, the model only has access to images from the %
allocentric views.
In the egocentric protocol,
to provide additional training images,
we allow models access to images from all views of the training split, but in evaluation, only egocentric images are used.
Further information can be found in \suppl.

\myparagraph{Metrics for \taskpose}
Our goal is to reconstruct the \threeD motion of the hands and an articulated object during dexterous manipulation from a video.
Importantly, our focus extends beyond hand-object poses and we require the reconstructed meshes to have accurate hand-object contact (CDev), and smooth motion (ACC).
Further, when a hand moves or articulates an object, vertices of the hand and the object in stable contact should move together (MDev).
To this end, we define the following metrics:
\begin{itemize}[noitemsep,topsep=0pt,parsep=0pt,partopsep=0pt]
\item \myparagraph{Contact Deviation (CDev)} 
For a frame, suppose $\{(\V{h}_i, \V{o}_i)\}_{i=1}^C$ are $C$ pairs of in-contact hand-object vertices ($<3mm$ distance in  \groundtruth), and $\{(\hat{\V{h}}_i, \hat{\V{o}}_i)\}_{i=1}^C$ are the corresponding predictions. CDev is defined as the average distance between $\hat{\V{h}}_i$ and $\hat{\V{o}}_i$ in millimeters:
\begin{align}
\resizebox{0.35\hsize}{!}{%
$
\frac{1}{C} \sum_{i=1}^C ||\hat{\V{h}}_i - \hat{\V{o}}_i||
$}
\end{align}
This metric reflects how much the hand vertices deviate from the supposed contact vertices on the object in the prediction.
\item \myparagraph{Motion Deviation (MDev)}
Given a \groundtruth sequence of a hand and an object, we denote vertex $i$ of the hand and vertex $j$ of the object at frame $t$ as $\V{h}_i^t$, $\V{o}_j^t$ respectively.
We use $(i, j, m, n)$ to denote $\V{h}_i^t$ has stable contact with $\V{o}_j^t$ during a window from frame $m$ to frame $n$, and they do not have contact at time $m-1$ and $n+1$ (\ie, longest contact window).
Hand-object vertex indices $(i, j)$ have stable contact in a window $(m,n)$ if they are close within a threshold $\alpha$ for every frame in the window:
\begin{align}
\resizebox{0.6\hsize}{!}{%
$
\forall t \in \{m, \cdots, n\}, \norm{\V{h}_i^t - \V{o}_j^t} \leq \alpha
$ .
}
\end{align}
Given the above definition, we extract a set of tuples $\{(i,j,m,n)\}$ from each GT sequence.
When two hand-object vertices $\V{h}_i^t$,  $\V{o}_j^t$ are in stable contact within a window, they should move in the same direction in consecutive frames.
To measure this, we define the motion deviation for a tuple $(i,j,m,n)$ of the predicted hand-object sequence $\hat{\V{h}}$ and $\hat{\V{o}}$ as
\begin{align}
\resizebox{0.5\hsize}{!}{%
$
 \frac{1}{n-m} \sum_{t=m+1}^{n} 
||\delta \hat{\V{h}}_i^t - \delta \hat{\V{o}}_j^t||
$
}
\end{align}
where $\delta \hat{\V{h}}_i^t = \hat{\V{h}}_i^t - \hat{\V{h}}_{i}^{t-1}$ and $\delta \hat{\V{o}}_j^t = \hat{\V{o}}_j^t - \hat{\V{o}}_{j}^{t-1}$.
Intuitively, this measures the disagreement in the moving direction between consecutive frames of in-contact hand-object vertices in the window $(m,n)$.
We only consider longer motions by using windows with at least 0.5 second or 15 frames (\ie, $n-m + 1 \geq 15$) and we choose $\alpha = 3mm$ to detect a sufficient number of windows.
We compute this metric for all detected windows and average over them.
\item \myparagraph{Acceleration Error (ACC)} Following\ccite{kocabas2020vibe}, we report acceleration error in $m/{s^2}$ to measure the smoothness of the reconstruction, calculated as the difference in acceleration between the ground-truth and predicted vertex sequences for each hand and the object.
We subtract the root for each entity before computing the acceleration\ccite{kocabas2020vibe}.
The root for the object is defined as the center of an object's base.
Note  that we report this error in $m/s^2$, while \ccite{kocabas2020vibe} reports $mm/s^2$.
See \suppl for more details.
\end{itemize}
Apart from motion and contact, we need metrics to measure hand and object poses, and their relative translations:
\begin{itemize}[noitemsep,topsep=0pt,parsep=0pt,partopsep=0pt]
\item \myparagraph{Mean Per-Joint Position Error (MPJPE)} the L2 distance $(mm)$ between the $21$ predicted and \groundtruth joints for each hand after subtracting its root.
\item \myparagraph{Average Articulation Error (AAE)} the average absolute error between the predicted degree of articulation and the \groundtruth. 
\item \myparagraph{Success Rate} Following\cite{stevvsivc2020spatial,zakharov2019dpod}, to measure object reconstruction quality, we use a success rate metric that is independent of the object size. 
It is the percentage of predicted object vertices having L2 error to the \groundtruth that is less than 5\% of the object diameter:
\begin{align}
\resizebox{0.7\hsize}{!}{%
$
    \frac{1}{V_o} \sum_{i=1}^{V_o} \mathds{1}(\norm{\V{o}_i - \hat{\V{o}}_i} < 0.05D) \times 100\%
$}
\end{align}
where $D$, $V_o$, $\V{o}_i$, $\V{\hat{o}}_i$ are the diameter, the number of object vertices, \groundtruth and predicted object vertices, and $\mathds{1}(\cdot)$ is the indicator function.
To decouple the effect of root estimation, we subtract the predicted and the \groundtruth vertices by their object roots respectively. 
The root is the center of each object's base.
\item \myparagraph{Mean Relative-Root Position Error (MRRPE)} 
Following~\cite{interhand,fan2021digit}, to measure the root translation of between hand-hand and hand-object, we use this metric to measure the relative root translation between two entities $a$ and $b$ in the scene,
\begin{equation}
    \resizebox{0.7\hsize}{!}{%
        $\operatorname{MRRPE}_{a \rightarrow b}=\left\|\left(\M{J}_{0}^{a}-\M{J}_{0}^{b}\right)-\left(\hat{\M{J}}_{0}^{a}-\hat{\M{J}}_{0}^{b}\right)\right\|_{2} \text{,}$%
        }
\end{equation}
where $a\in \{l, r, o\}$ and $b\in \{l, r, o\}$ and $l, r, o$ denote the left hand, right hand, and the object, 
$\M{J}_{0}\in \R^3$ is the \groundtruth root joint location and $\hat{\M{J}}_{0}$ the predicted one.
A graphical illustration of this metric can be found in \suppl.
\end{itemize}

\myparagraph{Metrics for \taskfield}
In this task, given images from a video, for each hand vertex $i$, we estimate its shortest distance $\mathbf{\hat{F}}_i^{r\rightarrow o}\in \R$ to the object (\ie, the distance field from a hand to the object) and vice versa.
Taking the field from the right hand to the object as an example, to quantify, we measure the average error between the predicted distances $\mathbf{\hat{F}}_i^{r\rightarrow o}$ and the \groundtruth distances $\mathbf{F}_i^{r\rightarrow o}$ in millimeters, which we call average distance error.
The error is computed as:
\begin{align}
\resizebox{0.4\hsize}{!}{%
$
    \frac{1}{V_r} \sum_{i=1}^{V_r} |\mathbf{F}_i^{r\rightarrow o} - \mathbf{\hat{F}}_i^{r\rightarrow o}|
$}
\end{align}
where $V_r$ is the number of right-hand vertices.
To measure smoothness, 
we estimate the distance field for every frame in each sequence.
We then compute the acceleration sequence for the predicted field sequence.
The acceleration error is computed as the average absolute difference between predicted and \groundtruth acceleration sequences.
See \suppl for the formula of acceleration error.

\section{Baselines and Experiments}

We present two tasks on \datasetname: \taskpose and \taskfield.
For \taskpose, we reconstruct the \threeD motion of two hands and an articulated object from a video.
For \taskfield, given a video, we estimate, for each hand vertex, the closest distance to the object and vice versa.
Here we detail and evaluate our baselines in the two tasks to lay the foundation for future comparison.

\subsection{\taskPose}

\myparagraph{Problem formulation}
Given a video, our goal is to reconstruct the \threeD motion of a subject's two hands and an articulated object in dexterous manipulation for every frame.
Our emphasis is to require the reconstructed hand-object meshes to be in temporally-consistent hand-object contact and motion during object articulation and manipulation.

\begin{figure}[]
    \centering
    \includegraphics[width=1.0\linewidth]{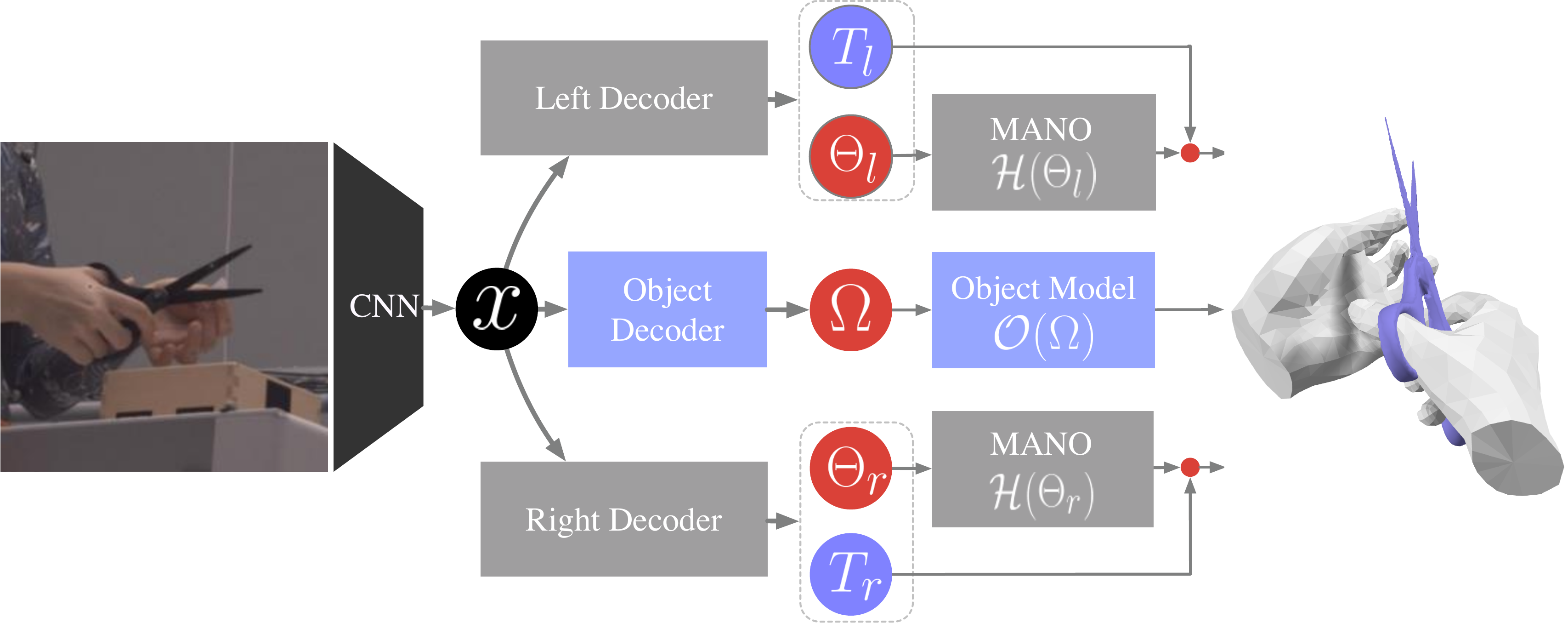}
    \caption{
    \subtitle{\methodnameSF architecture}
    The CNN encoder yields image features $x$.
    The hand decoders predict \mano parameters $\M{\Theta}_l, \M{\Theta}_r$ and their translation $\M{T}_l, \M{T}_r$ while the object decoder estimates the articulated object pose $\M{\Omega}$ consisting of the articulation, rotation and translation. 
    With parametric models of hands $\mathcal{H}(\M{\Theta})$ and articulated objects $\mathcal{O}(\M{\Omega})$, we obtain \threeD meshes for the two hands and the articulated object
    .}
    \label{fig:baseline}
    \vspace{-2mm}
\end{figure}

\begin{figure*}[t]
    \centerline{\includegraphics[width=1.0\linewidth]{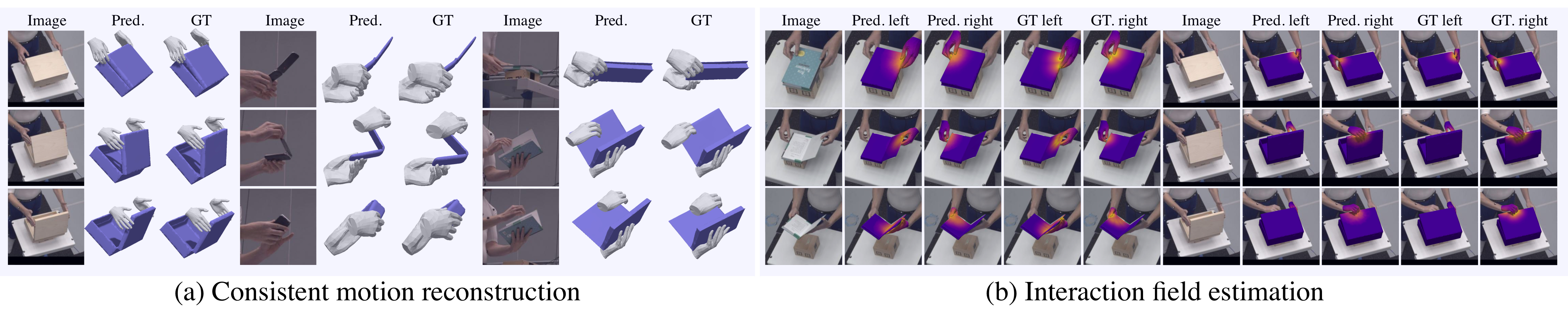}}
    \vspace{-2mm}
    \caption{
        \subtitle{Qualitative results of \methodnameLSTM (a) and \interfieldLSTM (b)} Best viewed in color and zoomed in. See \suppl for results of \methodnameSF and \interfieldSF.
    }
    \label{fig:quali_both}
\end{figure*}%

\begin{table*}[t]
\centering
\resizebox{1.0\linewidth}{!}{
\begin{tabular}{c|c|cc|cc|c|cc}
\toprule
                         &               & \multicolumn{2}{|c|}{Contact and Relative Position}   & \multicolumn{2}{|c|}{Motion}           & \multicolumn{1}{|c|}{Hand}  & \multicolumn{2}{c}{Object} \\\hline
Splits                   & Method        & CDev$_{ho}$ [$mm$] $\downarrow$& MRRPE$_{rl/ro}$ [$mm$] $\downarrow$ & MDev$_{ho}$ [$mm$] $\downarrow$ & ACC$_{h/o}$ [$m/{s^2}$] $\downarrow$ & MPJPE$_h$ [$mm$] $\downarrow$ & AAE [$^\circ$] $\downarrow$ & Success Rate [$\%$] $\uparrow$\\\hline
\multirow{2}{*}{Allo. Val}  & \methodnameSF  & 41.4 & 50.1/37.6 & 10.4 & 6.6/8.8 & 23.0  & 5.9     & 71.8          \\
                         & \methodnameLSTM & \textbf{38.8} & \textbf{47.1/36.8} & \textbf{8.9} & \textbf{5.6/6.9} & \textbf{22.9}  & \textbf{5.8}    & \textbf{74.9}          \\
\hline
\multirow{2}{*}{Allo. Test} & \methodnameSF  & 41.6 & 52.4/\textbf{37.5} & 10.4 & 5.7/7.6 & \textbf{21.5}  & 5.4      & 71.4          \\
                         & \methodnameLSTM & \textbf{38.9} & \textbf{49.2}/37.7 & \textbf{9.3} & \textbf{5.0/6.1} & \textbf{21.5}  & \textbf{5.2}      & \textbf{73.5}    \\
\hline
\multirow{2}{*}{Ego. Val}  & \methodnameSF  & \textbf{44.1} & \textbf{33.9/36.8} &  11.8 & 6.3/11.3 & \textbf{22.9}  & \textbf{8.0}     & 59.0         \\
                         & \methodnameLSTM & 44.5 & 39.3/39.0 & \textbf{8.1} &  \textbf{4.3/7.2} & 23.8  & \textbf{8.0}      & \textbf{59.1}        \\
\hline
\multirow{2}{*}{Ego. Test} & \methodnameSF  & 44.7 & \textbf{28.3}/36.2 & 11.8 & 5.0/9.1 & \textbf{19.2} & \textbf{6.4}     & \textbf{53.9}         \\
                         & \methodnameLSTM & \textbf{43.3} & 31.8/\textbf{35.0} & \textbf{8.6} & \textbf{3.5/5.7} & 20.0  & 6.6    & 53.5   \\
\bottomrule
\end{tabular}
}
\caption{
\subtitle{Comparison of two reconstruction baselines}
Contact and relative position metrics  measure hand-object contact and relative root position prediction.
Motion metrics reflect motions with temporally-consistent contact and smoothness.
Hand and object metrics show root-relative reconstruction error.
See \refsec{sec:eval_protocol} for metric details.
We use $l,r,o$ to denote the left, the right hand, and the object.
To simplify the results, we average left and right hand metrics into one hand (denoted by $h$).
For example, CDev$_{ho}$ is the contact deviation between a hand and the object averaged over the two hands; MRRPE$_{rl/ro}$ denotes MRRPE$_{r\rightarrow l}$ and MRRPE$_{r\rightarrow o}$ between the slash.
}
\label{tab:sota}
\end{table*}

\myparagraph{Parametric models}
For brevity, we use $l$, $r$, and $o$ to denote the left hand, the right hand and the object.
For hands, we use \mano\ccite{mano} to represent the hand pose and shape by $\V{\Theta} = \{ \V{\theta}, \V{\beta} \}$, which consists of parameters for the pose $\V{\theta} \in \R^{48}$ (with global orientation) and the shape $\V{\beta}\in \R^{10}$.
The \mano model maps $\V{\Theta}$ to a shaped and posed \threeD mesh $\mathcal{H}(\V{\theta}, \V{\beta})\in \R^{778\times 3}$.
The \threeD joint locations $\M{J}=W\mathcal{H}\in \R^{J\times 3}$ are obtained using a pre-trained linear regressor $W$.
For each object, we construct a \threeD model $\mathcal{O}(\cdot)$ using the scanned object mesh, the estimated rotation axis, and the marker-vertex correspondences estimated in \refsec{sec:acquisition}.
The function takes as inputs the articulated object pose, $\V{\Omega}$, and outputs a posed \threeD mesh, 
$\mathcal{O}(\V{\Omega})\in \R^{V\times 3}$
, where $V$ denotes the object's number of vertices.
The %
object pose, $\V{\Omega} \in \R^7$, consists of 
the \oneD rotation (radians) for articulation, $\omega \in \R$, and 
the \sixD object rigid pose, \ie, its rotation, $\V{R}_o \in \R^3$, and translation, $\V{T}_o \in \R^3$.

\myparagraph{Baselines}
We introduce \methodname to estimate the poses of the two hands and the articulated object from \rgb images. 
We benchmark two versions of \methodname: a single-frame model (\methodname-SF), and a model with a recurrent architecture (\methodname-LSTM).
The LSTM baseline is used to allow a joint reasoning of hand and articulated object motions.
\refFig{fig:baseline} summarizes the architecture of \methodname-SF. 
Inspired by Hasson \etal\ccite{Hasson2020photometric,hasson2019obman}, we use an encoder-decoder architecture.
In particular, the CNN encoder takes in the input image 
and produces image features $\V{x}$.
The features are used by the hand decoders to estimate the parameters for the left and right hands, $\V{\Theta}_l$ and $\V{\Theta}_r$, as well as the translations for the two hands, $\V{T}_l$ and $\V{T}_r$.
Similarly, the object decoder predicts the articulated object pose, $\V{\Omega}$.
We use axis-angle for rotation and use the weak perspective camera model to estimate the translations\ccite{Boukhayma2019,Kanazawa2018_hmr,Kocabas_PARE_2021,Sarandi20FG,Zhang2019endtoend}.
The \methodname-LSTM model has the same architecture as \methodname-SF, except that it has an LSTM network to aggregate image features from multiple frames before passing them to the regression heads.
We train the models with \groundtruth \threeD keypoints, 2D projected keypoints, and the parameters of the hand and the object models.
We show details of the model and the training procedure in \suppl.

\myparagraph{Results}
\refFig{fig:quali_both}\colorRef{a} visualizes the predictions of one of our baselines, \methodnameLSTM with~\cite{kaufmann_vechev_aitviewer_2022}.
To see qualitative results of \methodnameSF, we refer to the \suppl.
\refTab{tab:sota} shows the quantitative evaluation of the two baseline models on \datasetname. 
The results show that, overall, the \methodnameLSTM model has temporally more consistent contact (CDev), and motion (MDev) between the hands and objects.
Further, it has smoother motion (ACC).
This demonstrates that temporal modelling is important for spatio-temporally consistent hand-object motion and contact.
See \refsec{sec:eval_protocol} for metric details.

\subsection{\taskField}

Existing contact detection methods mainly focus on binary contact estimation~\cite{Yang_2021_CPF,grady2021contactopt}.
In two-handed dexterous interactions, hands can be near the object, but not always in contact.
We define a general task of \taskfield to capture the relative spatial relations between hands and the object even when not in contact.

\myparagraph{Problem formulation}
We define an interaction field $\boldsymbol{F}^{a \rightarrow b}\in \R^{V_a}$ as the distance to the closest vertex on the mesh $\M{M}_b$ for all vertices in mesh $\M{M}_a$ where $V_a$ (or $V_b$) is the number of vertices in mesh $\M{M}_a$ (or $\M{M}_b$).
Formally, 
\begin{equation}
\resizebox{0.8\hsize}{!}{%
$
\boldsymbol{F}^{a \rightarrow b}_i = \min_{\ 1 \leqslant j \leqslant V_b}|| \V{v}_i^a - \V{v}_j^b ||_2, \quad 1 \leqslant i \leqslant V_a
$}
\end{equation}
where $\V{v}_k^m \in \R^3$ represents the $k$-th vertex of mesh $\M{M}_m$.
We define our task to estimate the interaction fields $\boldsymbol{F}^{l \rightarrow o}$,  $\boldsymbol{F}^{r \rightarrow o}$, $\boldsymbol{F}^{o \rightarrow l}$, and $\boldsymbol{F}^{o \rightarrow r}$ for each image.
In other words, for each vertex of each hand we aim to infer the closest distance to the object and vice-versa.

\begin{figure}[H]
  \centerline{\includegraphics[width=1.0\linewidth]{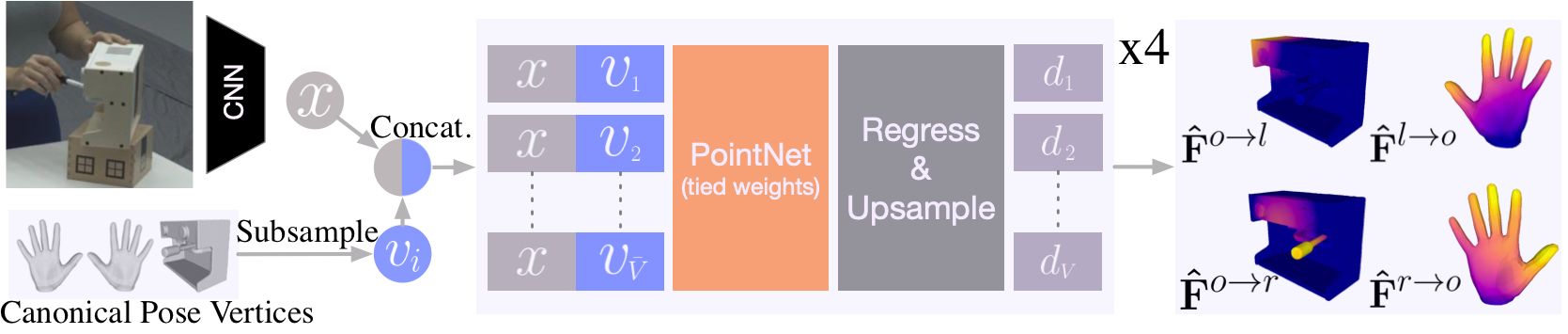}}
    \caption{
    \subtitle{\interfieldSF architecture}
    We concatenate image features $\V{x}$ to each subsampled hand-object vertex in canonical pose. The concatenated vectors are passed through a \pointnet and then regressed to distance values. The interaction field is visualized as a heatmap for each entity (bright: closest vertex is near).
    }
    \label{fig:interfield_est}
\end{figure}%

\myparagraph{Baselines}
We present \intermethod to estimate the interaction field from \rgb images. 
We benchmark two versions of \intermethod: a single-frame (\intermethod-SF) and a temporal baseline (\intermethod-LSTM).
The temporal model lets us evaluate the benefits of temporal information.
\refFig{fig:interfield_est} outlines the framework of \intermethod-SF.
Suppose that we estimate the field $\hat{\boldsymbol{F}}{}^{l \rightarrow o}$.
We first extract image features $\V{x} \in \R^d$ via a CNN backbone.
Next, we concatenate $\V{x}$ to each sub-sampled vertex of the left hand ($l$) in its canonical pose to obtain $\V{p}_i = [\V{x}; \V{v}_i] \in \R^{d+3}$ for all $1 \leqslant i \leqslant \bar{V}_l$ where $\bar{V_l}$ denotes the number of subsampled vertices.
All points $\V{p}_i$ are fed to a \pointnet \cite{qi2016pointnet} followed by a regression head that 
estimates the distance.
The predicted distances are upsampled to the full mesh.
For efficiency, we use subsampled vertices for the \pointnet and upsample for regression.
The remaining interaction fields are estimated via the same network with a shared CNN and \pointnet but 
different heads.
\intermethod-LSTM follows the same formulation except it has an LSTM to aggregate image features in a temporal window to jointly reason about hand-object motion.
See more training and baseline details in \suppl.

\smallskip

\begin{table}[t]
\centering
\resizebox{1.0\linewidth}{!}{
\begin{tabular}{c|c|c|c}
\toprule
Splits                       & Method          & Average Distance Error  [$mm$]$\downarrow$  & ACC [$m/{s^2}$]$\downarrow$\\
\hline
                             & \interfieldSF      & 9.6/9.9 & 3.0/2.9 \\
\multirow{-2}{*}{Allo. Val}  & \interfieldLSTM & \textbf{9.0/8.9} & \textbf{2.1/2.0} \\
\hline
                             & \interfieldSF      & 9.0/10.0 & 2.7/2.7\\
\multirow{-2}{*}{Allo. Test} & \interfieldLSTM & \textbf{8.7/9.1} & \textbf{1.9/1.9} \\
\hline
                             & \interfieldSF      & 8.8/9.2 & 2.4/2.3 \\
\multirow{-2}{*}{Ego. Val}   & \interfieldLSTM & \textbf{8.4/8.9} & \textbf{2.1/2.0} \\
\hline
                             & \interfieldSF      & 8.2/9.2 & 2.1/2.0 \\
\multirow{-2}{*}{Ego. Test}  & \interfieldLSTM & \textbf{8.0/9.1} & \textbf{1.8/1.8} \\
\bottomrule
\end{tabular}
}
\caption{
\subtitle{Comparison of two field estimation baselines}
To simplify the evaluation, we average metrics for the two hands into one.
The slashes denote the average distance error and the acceleration error for hand-to-object/object-to-hand.
}
\label{tab:sota_field}
\end{table}

\myparagraph{Results}
\refFig{fig:quali_both}\colorRef{b} shows qualitative samples of \intermethod-LSTM.
The predicted values are visualized as heatmaps over the meshes of the respective hands or objects.
A ``hotter" region denotes closer distances.
Note that the \groundtruth meshes are only used for visualization; they are not network inputs.
We find that the predicted fields correlate well with the ground truth.
\refTab{tab:sota_field} shows the performance of our baselines. 
The results show that modeling the hand-object interaction field over time yields more accurate results (see distance error), and smoother predictions (ACC).

\section{Conclusions}\label{secconclusion}
We introduce \datasetname, the first dataset with two hands dexterously manipulating articulated objects that
includes high-quality \threeD~\groundtruth for hands, and objects
together with synchronized video.
\datasetname has  a total of \numImages \rgb images from \numAllo static views and \numEgo egocentric view of \numSubs subjects interacting with \numObjs articulated objects.
We present two tasks on \datasetname. First is \textit{\taskpose}. Given a video, we  reconstruct two hands and an articulated object in \threeD for every frame, such that their motions are spatio-temporally consistent. The second task is \textit{\taskfield}, where we estimate dense relative hand-object distances from images in a video.
We present two baselines \methodname and \intermethod for the two tasks respectively, and evaluate them on \datasetname to lay the foundation for future work.

\myparagraph{Future directions} 
\datasetname can enable a range of  tasks related to hand manipulation with object articulation. 
First, methods for generating hand-object interaction focus on generating grasps of rigid objects~\cite{karunratanakul2020graspField,christen2022dgrasp}, but less work has been done on generating dexterous bimanual manipulation motion with objects\cite{zhang2021manipnet,chen2022towards} and prior work does not generate interaction with articulated objects (\eg, ``cutting with scissors").
\datasetname can enable these new generation tasks, and extend them to the full-body~\cite{taheri2021goal} with our \smplx\ \groundtruth.
Second, we introduce tasks of \taskpose and \taskfield. 
Future work could leverage the interaction field representation for pose estimation to improve hand-object contact in reconstruction.
Finally, articulated object pose estimators~\cite{Li2020categoryaArticulated} from depth images do not consider humans in the scene.
The rendered depth images in \datasetname can be used to benchmark such methods in more realistic settings.

\smallskip
{
\myparagraph{Acknowledgements} 
The authors deeply thank 
Tsvetelina Alexiadis (TA) for trial coordination; 
Markus H{\"o}schle (MH), Senya Polikovsky, Matvey Safroshkin, Tobias Bauch (TB) for the capture setup; 
MH, TA, Galina Henz for data capture; Giorgio Becherini, Nima Ghorbani for \mosh;
Priyanka Patel for alignment;
Leyre S{\'a}nchez Vinuela, Andres Camilo Mendoza Patino, Mustafa Alperen Ekinci for data cleaning;
TB for Vicon support;
MH, Jakob Reinhardt for object scanning;
Taylor McConnell for Vicon support, and data cleaning coordination;
Benjamin Pellkofer for IT support.
We also thank Xu Chen, Adrian Spurr, Jie Song for insightful discussion.
OT and DT were partially funded by the German Federal
Ministry of Education and Research (BMBF): T{\"u}bingen AI Center, FKZ: 01IS18039B.
DT's work was partially performed at the MPI-IS.

\myparagraph{Disclosure}
\url{https://files.is.tue.mpg.de/black/CoI_CVPR_2023.txt}
}

{\small
\balance
\bibliographystyle{ieee_fullname}
\bibliography{0_config/egbib}

\begin{thebibliography}{10}\itemsep=-1pt

\bibitem{ballan2012motion}
Luca Ballan, Aparna Taneja, J{\"u}rgen Gall, Luc Van~Gool, and Marc Pollefeys.
\newblock Motion capture of hands in action using discriminative salient
  points.
\newblock In {\em {European Conference on Computer Vision (ECCV)}}, pages
  640--653, 2012.

\bibitem{dillmann2005grasp}
Keni Bernardin, Koichi Ogawara, Katsushi Ikeuchi, and Ruediger Dillmann.
\newblock A sensor fusion approach for recognizing continuous human grasping
  sequences using hidden markov models.
\newblock {\em Transactions on Robotics}, 21(1):47--57, 2005.

\bibitem{bhatnagar2022behave}
Bharat~Lal Bhatnagar, Xianghui Xie, Ilya~A Petrov, Cristian Sminchisescu,
  Christian Theobalt, and Gerard Pons-Moll.
\newblock {BEHAVE}: Dataset and method for tracking human object interactions.
\newblock In {\em {Computer Vision and Pattern Recognition (CVPR)}}, pages
  15935--15946, 2022.

\bibitem{BEDLAM:CVPR:2023}
Michael~J. Black, Priyanka Patel, Joachim Tesch, and Jinlong Yang.
\newblock {BEDLAM}: A dataset of bodies exhibiting detailed lifelike animated
  motion.
\newblock In {\em {Computer Vision and Pattern Recognition (CVPR)}}, June 2023.

\bibitem{Boukhayma2019}
Adnane Boukhayma, Rodrigo de Bem, and Philip H.~S. Torr.
\newblock {3D} hand shape and pose from images in the wild.
\newblock In {\em {Computer Vision and Pattern Recognition (CVPR)}}, pages
  10843--10852, 2019.

\bibitem{contactpose_2020}
Samarth Brahmbhatt, Chengcheng Tang, Christopher~D. Twigg, Charles~C. Kemp, and
  James Hays.
\newblock {ContactPose}: {A} dataset of grasps with object contact and hand
  pose.
\newblock In {\em {European Conference on Computer Vision (ECCV)}}, volume
  12358, pages 361--378, 2020.

\bibitem{cao2021handobject}
Zhe Cao, Ilija Radosavovic, Angjoo Kanazawa, and Jitendra Malik.
\newblock Reconstructing hand-object interactions in the wild.
\newblock In {\em {International Conference on Computer Vision ({ICCV})}},
  pages 12417--12426, 2021.

\bibitem{dexycb}
Yu-Wei Chao, Wei Yang, Yu Xiang, Pavlo Molchanov, Ankur Handa, Jonathan
  Tremblay, Yashraj~S. Narang, Karl {Van Wyk}, Umar Iqbal, Stan Birchfield, Jan
  Kautz, and Dieter Fox.
\newblock {DexYCB}: {A} benchmark for capturing hand grasping of objects.
\newblock In {\em {Computer Vision and Pattern Recognition (CVPR)}}, pages
  9044--9053, 2021.

\bibitem{chen2023hot}
Yixin Chen, Sai~Kumar Dwivedi, Michael~J. Black, and Dimitrios Tzionas.
\newblock Detecting human-object contact in images.
\newblock In {\em {Computer Vision and Pattern Recognition (CVPR)}}, June 2023.

\bibitem{chen2022towards}
Yuanpei Chen, Yaodong Yang, Tianhao Wu, Shengjie Wang, Xidong Feng, Jiechuang
  Jiang, Stephen~Marcus McAleer, Hao Dong, Zongqing Lu, and Song-Chun Zhu.
\newblock Towards human-level bimanual dexterous manipulation with
  reinforcement learning.
\newblock {\em arXiv preprint arXiv:2206.08686}, 2022.

\bibitem{christen2022dgrasp}
Sammy Christen, Muhammed Kocabas, Emre Aksan, Jemin Hwangbo, Jie Song, and
  Otmar Hilliges.
\newblock {D-Grasp}: {P}hysically plausible dynamic grasp synthesis for
  hand-object interactions.
\newblock In {\em {Computer Vision and Pattern Recognition (CVPR)}}, pages
  20545--20554, 2022.

\bibitem{corona2020ganhand}
Enric Corona, Albert Pumarola, Guillem Aleny{\`{a}}, Francesc Moreno{-}Noguer,
  and Gr{\'{e}}gory Rogez.
\newblock {GanHand}: {P}redicting human grasp affordances in multi-object
  scenes.
\newblock In {\em {Computer Vision and Pattern Recognition (CVPR)}}, pages
  5030--5040, 2020.

\bibitem{deng2009imagenet}
Jia Deng, Wei Dong, Richard Socher, Li-Jia Li, Kai Li, and Li Fei-Fei.
\newblock Imagenet: A large-scale hierarchical image database.
\newblock In {\em {Computer Vision and Pattern Recognition (CVPR)}}, pages
  248--255. Ieee, 2009.

\bibitem{Doosti2020hopeNet}
Bardia Doosti, Shujon Naha, Majid Mirbagheri, and David~J. Crandall.
\newblock {HOPE-Net}: {A} graph-based model for hand-object pose estimation.
\newblock In {\em {Computer Vision and Pattern Recognition (CVPR)}}, pages
  6607--6616, 2020.

\bibitem{eldar1994fps}
Yuval Eldar, Michael Lindenbaum, Moshe Porat, and Yehoshua~Y. Zeevi.
\newblock The farthest point strategy for progressive image sampling.
\newblock In {\em {International Conference on Pattern Recognition (ICPR)}},
  pages 93--97, 1994.

\bibitem{eldar1997fps}
Yuval Eldar, Michael Lindenbaum, Moshe Porat, and Yehoshua~Y. Zeevi.
\newblock The farthest point strategy for progressive image sampling.
\newblock {\em {Transactions on Image Processing (TIP)}}, 6(9):1305--1315,
  1997.

\bibitem{fan2021digit}
Zicong Fan, Adrian Spurr, Muhammed Kocabas, Siyu Tang, Michael~J. Black, and
  Otmar Hilliges.
\newblock Learning to disambiguate strongly interacting hands via probabilistic
  per-pixel part segmentation.
\newblock In {\em {International Conference on 3D Vision (3DV)}}, pages 1--10,
  2021.

\bibitem{Feix_GRASP_2016}
Thomas Feix, Javier Romero, Heinz-Bodo Schmiedmayer, Aaron~M. Dollar, and
  Danica Kragic.
\newblock The grasp taxonomy of human grasp types.
\newblock {\em {Transactions on Human-Machine Systems (THMS)}}, 46(1):66--77,
  2016.

\bibitem{FirstPersonAction_CVPR2018}
Guillermo Garcia-Hernando, Shanxin Yuan, Seungryul Baek, and Tae-Kyun Kim.
\newblock First-person hand action benchmark with {RGB-D} videos and {3D} hand
  pose annotations.
\newblock In {\em {Computer Vision and Pattern Recognition (CVPR)}}, pages
  409--419, 2018.

\bibitem{grady2022pressurevision}
Patrick Grady, Chengcheng Tang, Samarth Brahmbhatt, Christopher~D. Twigg,
  Chengde Wan, James Hays, and Charles~C. Kemp.
\newblock {PressureVision}: {E}stimating hand pressure from a single {RGB}
  image.
\newblock {\em {European Conference on Computer Vision (ECCV)}},
  13666:328--345, 2022.

\bibitem{grady2021contactopt}
Patrick Grady, Chengcheng Tang, Christopher~D. Twigg, Minh Vo, Samarth
  Brahmbhatt, and Charles~C. Kemp.
\newblock {ContactOpt}: {O}ptimizing contact to improve grasps.
\newblock In {\em {Computer Vision and Pattern Recognition (CVPR)}}, pages
  1471--1481, 2021.

\bibitem{hampali2020honnotate}
Shreyas Hampali, Mahdi Rad, Markus Oberweger, and Vincent Lepetit.
\newblock {HOnnotate}: {A} method for {3D} annotation of hand and object poses.
\newblock In {\em {Computer Vision and Pattern Recognition (CVPR)}}, pages
  3193--3203, 2020.

\bibitem{hampali2022keypoint}
Shreyas Hampali, Sayan~Deb Sarkar, Mahdi Rad, and Vincent Lepetit.
\newblock Keypoint transformer: {S}olving joint identification in challenging
  hands and object interactions for accurate {3D} pose estimation.
\newblock In {\em {Computer Vision and Pattern Recognition (CVPR)}}, pages
  11090--11100, 2022.

\bibitem{Hasson2020photometric}
Yana Hasson, Bugra Tekin, Federica Bogo, Ivan Laptev, Marc Pollefeys, and
  Cordelia Schmid.
\newblock Leveraging photometric consistency over time for sparsely supervised
  hand-object reconstruction.
\newblock In {\em {Computer Vision and Pattern Recognition (CVPR)}}, pages
  568--577, 2020.

\bibitem{hasson2019obman}
Yana Hasson, G{\"{u}}l Varol, Dimitrios Tzionas, Igor Kalevatykh, Michael~J.
  Black, Ivan Laptev, and Cordelia Schmid.
\newblock Learning joint reconstruction of hands and manipulated objects.
\newblock In {\em {Computer Vision and Pattern Recognition (CVPR)}}, pages
  11807--11816, 2019.

\bibitem{huang2022rich}
Chun-Hao~P. Huang, Hongwei Yi, Markus H{\"o}schle, Matvey Safroshkin,
  Tsvetelina Alexiadis, Senya Polikovsky, Daniel Scharstein, and Michael~J.
  Black.
\newblock Capturing and inferring dense full-body human-scene contact.
\newblock In {\em {Computer Vision and Pattern Recognition (CVPR)}}, pages
  13274--13285, June 2022.

\bibitem{huang2022intercap}
Yinghao Huang, Omid Taheri, Michael~J. Black, and Dimitrios Tzionas.
\newblock {InterCap}: {J}oint markerless {3D} tracking of humans and objects in
  interaction.
\newblock In {\em {German Conference on Pattern Recognition (GCPR)}}, volume
  13485, pages 281--299, 2022.

\bibitem{iqbal2018hand}
Umar Iqbal, Pavlo Molchanov, Thomas Breuel~Juergen Gall, and Jan Kautz.
\newblock Hand pose estimation via latent {2.5D} heatmap regression.
\newblock In {\em {European Conference on Computer Vision (ECCV)}}, pages
  118--134, 2018.

\bibitem{kamakura1980grasp}
Noriko Kamakura, Michiko Matsuo, Harumi Ishii, Fumiko Mitsuboshi, and Yoriko
  Miura.
\newblock Patterns of static prehension in normal hands.
\newblock {\em American Journal of Occupational Therapy}, 34(7):437--445, 1980.

\bibitem{Kanazawa2018_hmr}
Angjoo Kanazawa, Michael~J. Black, David~W. Jacobs, and Jitendra Malik.
\newblock End-to-end recovery of human shape and pose.
\newblock In {\em {Computer Vision and Pattern Recognition (CVPR)}}, pages
  7122--7131, 2018.

\bibitem{karunratanakul2020graspField}
Korrawe Karunratanakul, Jinlong Yang, Yan Zhang, Michael~J. Black, Krikamol
  Muandet, and Siyu Tang.
\newblock {Grasping Field}: {L}earning implicit representations for human
  grasps.
\newblock In {\em {International Conference on 3D Vision (3DV)}}, pages
  333--344, 2020.

\bibitem{kaufmann_vechev_aitviewer_2022}
Manuel Kaufmann, Velko Vechev, and Dario Mylonopoulos.
\newblock {aitviewer}, 7 2022.

\bibitem{kingma2014adam}
Diederik~P. Kingma and Jimmy Ba.
\newblock Adam: {A} method for stochastic optimization.
\newblock In {\em {International Conference on Learning Representations
  (ICLR)}}, 2015.

\bibitem{kocabas2020vibe}
Muhammed Kocabas, Nikos Athanasiou, and Michael~J. Black.
\newblock {VIBE}: Video inference for human body pose and shape estimation.
\newblock In {\em {Computer Vision and Pattern Recognition (CVPR)}}, pages
  5253--5263, 2020.

\bibitem{Kocabas_PARE_2021}
Muhammed Kocabas, Chun-Hao~P. Huang, Otmar Hilliges, and Michael~J. Black.
\newblock {PARE}: {P}art attention regressor for {3D} human body estimation.
\newblock In {\em {International Conference on Computer Vision ({ICCV})}},
  pages 11127--11137, 2021.

\bibitem{mkocabas2021spec}
Muhammed Kocabas, Chun-Hao~P. Huang, Joachim Tesch, Lea M\"uller, Otmar
  Hilliges, and Michael~J. Black.
\newblock {SPEC}: Seeing people in the wild with an estimated camera.
\newblock In {\em {International Conference on Computer Vision ({ICCV})}},
  pages 11035--11045, 2021.

\bibitem{kwon2021h2o}
Taein Kwon, Bugra Tekin, Jan St{\"u}hmer, Federica Bogo, and Marc Pollefeys.
\newblock {H2O}: {T}wo hands manipulating objects for first person interaction
  recognition.
\newblock In {\em {International Conference on Computer Vision ({ICCV})}},
  pages 10138--10148, 2021.

\bibitem{li2022interacting}
Mengcheng Li, Liang An, Hongwen Zhang, Lianpeng Wu, Feng Chen, Tao Yu, and
  Yebin Liu.
\newblock Interacting attention graph for single image two-hand reconstruction.
\newblock In {\em {Computer Vision and Pattern Recognition (CVPR)}}, pages
  2761--2770, 2022.

\bibitem{Li2020categoryaArticulated}
Xiaolong Li, He Wang, Li Yi, Leonidas~J. Guibas, A.~Lynn Abbott, and Shuran
  Song.
\newblock Category-level articulated object pose estimation.
\newblock In {\em {Computer Vision and Pattern Recognition (CVPR)}}, pages
  3703--3712, 2020.

\bibitem{liu2021semi}
Shaowei Liu, Hanwen Jiang, Jiarui Xu, Sifei Liu, and Xiaolong Wang.
\newblock Semi-supervised {3D} hand-object poses estimation with interactions
  in time.
\newblock In {\em {Computer Vision and Pattern Recognition (CVPR)}}, pages
  14687--14697, 2021.

\bibitem{liu2022hoi4d}
Yunze Liu, Yun Liu, Che Jiang, Kangbo Lyu, Weikang Wan, Hao Shen, Boqiang
  Liang, Zhoujie Fu, He Wang, and Li Yi.
\newblock {HOI4D}: A {4D} egocentric dataset for category-level human-object
  interaction.
\newblock In {\em {Computer Vision and Pattern Recognition (CVPR)}}, pages
  21013--21022, 2022.

\bibitem{AMASS_2019}
Naureen Mahmood, Nima Ghorbani, Nikolaus F.~Troje, Gerard Pons-Moll, and
  Michael~J. Black.
\newblock {AMASS}: {A}rchive of motion capture as surface shapes.
\newblock In {\em {International Conference on Computer Vision ({ICCV})}},
  pages 5441--5450, 2019.

\bibitem{michel2015pose}
Frank Michel, Alexander Krull, Eric Brachmann, Michael~Ying Yang, Stefan
  Gumhold, and Carsten Rother.
\newblock Pose estimation of kinematic chain instances via object coordinate
  regression.
\newblock In {\em {British Machine Vision Conference (BMVC)}}, pages
  181.1--181.11, 2015.

\bibitem{interhand}
Gyeongsik Moon, Shoou-I Yu, He Wen, Takaaki Shiratori, and Kyoung~Mu Lee.
\newblock {InterHand2.6M}: {A} dataset and baseline for {3D} interacting hand
  pose estimation from a single {RGB} image.
\newblock In {\em {European Conference on Computer Vision (ECCV)}}, volume
  12365, pages 548--564, 2020.

\bibitem{Mueller2018ganerated}
Franziska Mueller, Florian Bernard, Oleksandr Sotnychenko, Dushyant Mehta,
  Srinath Sridhar, Dan Casas, and Christian Theobalt.
\newblock {GANerated} hands for real-time {3D} hand tracking from monocular
  {RGB}.
\newblock In {\em {Computer Vision and Pattern Recognition (CVPR)}}, pages
  49--59, 2018.

\bibitem{mueller2017realtime}
Franziska Mueller, Dushyant Mehta, Oleksandr Sotnychenko, Srinath Sridhar, Dan
  Casas, and Christian Theobalt.
\newblock Real-time hand tracking under occlusion from an egocentric {RGB-D}
  sensor.
\newblock In {\em {International Conference on Computer Vision ({ICCV})}},
  pages 1163--1172, 2017.

\bibitem{narasimhaswamy2020detecting}
Supreeth Narasimhaswamy, Trung Nguyen, and Minh~Hoai Nguyen.
\newblock Detecting hands and recognizing physical contact in the wild.
\newblock In {\em {Conference on Neural Information Processing Systems
  (NeurIPS)}}, volume~33, 2020.

\bibitem{pavlakos2019expressive}
Georgios Pavlakos, Vasileios Choutas, Nima Ghorbani, Timo Bolkart, Ahmed A.~A.
  Osman, Dimitrios Tzionas, and Michael~J. Black.
\newblock Expressive body capture: {3D} hands, face, and body from a single
  image.
\newblock In {\em {Computer Vision and Pattern Recognition (CVPR)}}, pages
  10975--10985, 2019.

\bibitem{pham2018pami}
Tu{-}Hoa Pham, Nikolaos Kyriazis, Antonis~A. Argyros, and Abderrahmane Kheddar.
\newblock Hand-object contact force estimation from markerless visual tracking.
\newblock {\em {Transactions on Pattern Analysis and Machine Intelligence
  (TPAMI)}}, 40(12):2883--2896, 2018.

\bibitem{qi2016pointnet}
Charles~R. Qi, Hao Su, Kaichun Mo, and Leonidas~J. Guibas.
\newblock {PointNet}: {D}eep learning on point sets for {3D} classification and
  segmentation.
\newblock In {\em {Computer Vision and Pattern Recognition (CVPR)}}, pages
  77--85, 2017.

\bibitem{Rehg:1994}
James~M. Rehg and Takeo Kanade.
\newblock Visual tracking of high {DOF} articulated structures: An application
  to human hand tracking.
\newblock In {\em {European Conference on Computer Vision (ECCV)}}, volume 801,
  pages 35--46, 1994.

\bibitem{Rogez2015everyday}
Gr{\'{e}}gory Rogez, James~Steven {Supan{\v{c}}i{\v{c}} III}, and Deva Ramanan.
\newblock Understanding everyday hands in action from {RGB-D} images.
\newblock In {\em {International Conference on Computer Vision ({ICCV})}},
  pages 3889--3897, 2015.

\bibitem{mano}
Javier Romero, Dimitrios Tzionas, and Michael~J. Black.
\newblock Embodied hands: {M}odeling and capturing hands and bodies together.
\newblock {\em {Transactions on Graphics (TOG)}}, 36(6):245:1--245:17, 2017.

\bibitem{Sarandi20FG}
Istv\'an S\'ar\'andi, Timm Linder, Kai~O. Arras, and Bastian Leibe.
\newblock Metric-scale truncation-robust heatmaps for {3D} human pose
  estimation.
\newblock In {\em {International Conference on Automatic Face \& Gesture
  Recognition (FG)}}, pages 407--414, 2020.

\bibitem{sener2022assembly101}
Fadime Sener, Dibyadip Chatterjee, Daniel Shelepov, Kun He, Dipika Singhania,
  Robert Wang, and Angela Yao.
\newblock {Assembly101}: A large-scale multi-view video dataset for
  understanding procedural activities.
\newblock In {\em {Computer Vision and Pattern Recognition (CVPR)}}, pages
  21064--21074, 2022.

\bibitem{shan2020contact}
Dandan Shan, Jiaqi Geng, Michelle Shu, and David~F. Fouhey.
\newblock Understanding human hands in contact at internet scale.
\newblock In {\em {Computer Vision and Pattern Recognition (CVPR)}}, pages
  9866--9875, 2020.

\bibitem{simon2017hand}
Tomas Simon, Hanbyul Joo, Iain Matthews, and Yaser Sheikh.
\newblock Hand keypoint detection in single images using multiview
  bootstrapping.
\newblock In {\em {Computer Vision and Pattern Recognition (CVPR)}}, pages
  4645--4653, 2017.

\bibitem{spurr2021peclr}
Adrian Spurr, Aneesh Dahiya, Xi Wang, Xucong Zhang, and Otmar Hilliges.
\newblock Self-supervised {3D} hand pose estimation from monocular {RGB} via
  contrastive learning.
\newblock In {\em {International Conference on Computer Vision ({ICCV})}},
  pages 11210--11219, 2021.

\bibitem{spurr2020eccv}
Adrian Spurr, Umar Iqbal, Pavlo Molchanov, Otmar Hilliges, and Jan Kautz.
\newblock Weakly supervised {3D} hand pose estimation via biomechanical
  constraints.
\newblock In {\em {European Conference on Computer Vision (ECCV)}}, volume
  12362, pages 211--228, 2020.

\bibitem{Spurr2018crossmodal}
Adrian Spurr, Jie Song, Seonwook Park, and Otmar Hilliges.
\newblock Cross-modal deep variational hand pose estimation.
\newblock In {\em {Computer Vision and Pattern Recognition (CVPR)}}, pages
  89--98, 2018.

\bibitem{RealtimeHO_ECCV2016}
Srinath Sridhar, Franziska Mueller, Michael Zollhoefer, Dan Casas, Antti
  Oulasvirta, and Christian Theobalt.
\newblock Real-time joint tracking of a hand manipulating an object from
  {RGB-D} input.
\newblock In {\em {European Conference on Computer Vision (ECCV)}}, volume
  9906, pages 294--310, 2016.

\bibitem{stevvsivc2020spatial}
Stefan Stev{\v{s}}i{\v{c}} and Otmar Hilliges.
\newblock Spatial attention improves iterative {6D} object pose estimation.
\newblock In {\em {International Conference on 3D Vision (3DV)}}, pages
  1070--1078, 2020.

\bibitem{sun2019deep}
Ke Sun, Bin Xiao, Dong Liu, and Jingdong Wang.
\newblock Deep high-resolution representation learning for human pose
  estimation.
\newblock In {\em {Computer Vision and Pattern Recognition (CVPR)}}, 2019.

\bibitem{taheri2021goal}
Omid Taheri, Vassileios Choutas, Michael~J. Black, and Dimitrios Tzionas.
\newblock {GOAL}: {G}enerating {4D} whole-body motion for hand-object grasping.
\newblock In {\em {Computer Vision and Pattern Recognition (CVPR)}}, pages
  13253--13263, 2022.

\bibitem{grab}
Omid Taheri, Nima Ghorbani, Michael~J. Black, and Dimitrios Tzionas.
\newblock {GRAB}: {A} dataset of whole-body human grasping of objects.
\newblock In {\em {European Conference on Computer Vision (ECCV)}}, volume
  12349, pages 581--600, 2020.

\bibitem{tekin2019ho}
Bugra Tekin, Federica Bogo, and Marc Pollefeys.
\newblock {H+O}: Unified egocentric recognition of {3D} hand-object poses and
  interactions.
\newblock In {\em {Computer Vision and Pattern Recognition (CVPR)}}, pages
  4511--4520, 2019.

\bibitem{3dmdhand}
{3dMDhand} system series.
\newblock \url{https://3dmd.com/products/}.

\bibitem{tripathi2023ipman}
Shashank Tripathi, Lea M{\"u}ller, Chun-Hao~P. Huang, Taheri Omid, Michael~J.
  Black, and Dimitrios Tzionas.
\newblock {3D} human pose estimation via intuitive physics.
\newblock In {\em {Computer Vision and Pattern Recognition (CVPR)}}, June 2023.

\bibitem{Tsoli2018deformable}
Aggeliki Tsoli and Antonis~A. Argyros.
\newblock Joint {3D} tracking of a deformable object in interaction with a
  hand.
\newblock In {\em {European Conference on Computer Vision (ECCV)}}, volume
  11218, pages 504--520, 2018.

\bibitem{tzionas2016capturing}
Dimitrios Tzionas, Luca Ballan, Abhilash Srikantha, Pablo Aponte, Marc
  Pollefeys, and Juergen Gall.
\newblock Capturing hands in action using discriminative salient points and
  physics simulation.
\newblock {\em {International Journal of Computer Vision (IJCV)}},
  118(2):172--193, 2016.

\bibitem{tzionas2013directional}
Dimitrios Tzionas and Juergen Gall.
\newblock A comparison of directional distances for hand pose estimation.
\newblock In {\em {German Conference on Pattern Recognition (GCPR)}}, volume
  8142, pages 131--141, 2013.

\bibitem{Tzionas2015inhand}
Dimitrios Tzionas and Juergen Gall.
\newblock {3D} object reconstruction from hand-object interactions.
\newblock In {\em {International Conference on Computer Vision ({ICCV})}},
  pages 729--737, 2015.

\bibitem{tzionas2016articulated}
Dimitrios Tzionas and Juergen Gall.
\newblock Reconstructing articulated rigged models from {RGB-D} videos.
\newblock In {\em {European Conference on Computer Vision Workshops (ECCVw)}},
  volume 9915, pages 620--633, 2016.

\bibitem{JMLR:v9:vandermaaten08a}
Laurens van~der Maaten and Geoffrey Hinton.
\newblock Visualizing data using {t-SNE}.
\newblock {\em {Journal of Machine Learning Research (JMLR)}},
  9(86):2579--2605, 2008.

\bibitem{viconVantageWEB}
{Vicon Vantage}: Cutting edge, flagship camera with intelligent feedback and
  resolution.
\newblock \url{https://www.vicon.com/hardware/cameras/vantage}.

\bibitem{xiang2020sapien}
Fanbo Xiang, Yuzhe Qin, Kaichun Mo, Yikuan Xia, Hao Zhu, Fangchen Liu, Minghua
  Liu, Hanxiao Jiang, Yifu Yuan, He Wang, Li Yi, Angel~X. Chang, Leonidas~J.
  Guibas, and Hao Su.
\newblock {SAPIEN}: {A} simulated part-based interactive environment.
\newblock In {\em {Computer Vision and Pattern Recognition (CVPR)}}, pages
  11094--11104, 2020.

\bibitem{Yang_2021_CPF}
Lixin Yang, Xinyu Zhan, Kailin Li, Wenqiang Xu, Jiefeng Li, and Cewu Lu.
\newblock {CPF}: {L}earning a contact potential field to model the hand-object
  interaction.
\newblock In {\em {International Conference on Computer Vision ({ICCV})}},
  pages 11097--11106, 2021.

\bibitem{yin2023hi4d}
Yifei Yin, Chen Guo, Manuel Kaufmann, Juan Zarate, Jie Song, and Otmar
  Hilliges.
\newblock {Hi4D}: {4D} instance segmentation of close human interaction.
\newblock In {\em {Computer Vision and Pattern Recognition (CVPR)}}, 2023.

\bibitem{zakharov2019dpod}
Sergey Zakharov, Ivan Shugurov, and Slobodan Ilic.
\newblock {DPOD}: {6D} pose object detector and refiner.
\newblock In {\em {International Conference on Computer Vision ({ICCV})}},
  pages 1941--1950, 2019.

\bibitem{zhang2021interacting}
Baowen Zhang, Yangang Wang, Xiaoming Deng, Yinda Zhang, Ping Tan, Cuixia Ma,
  and Hongan Wang.
\newblock Interacting two-hand {3D} pose and shape reconstruction from single
  color image.
\newblock In {\em {International Conference on Computer Vision ({ICCV})}},
  pages 11354--11363, 2021.

\bibitem{zhang2021manipnet}
He Zhang, Yuting Ye, Takaaki Shiratori, and Taku Komura.
\newblock {ManipNet}: {N}eural manipulation synthesis with a hand-object
  spatial representation.
\newblock {\em {Transactions on Graphics (TOG)}}, 40(4):1--14, 2021.

\bibitem{zhang2021inhand}
Hao Zhang, Yuxiao Zhou, Yifei Tian, Jun{-}Hai Yong, and Feng Xu.
\newblock Single depth view based real-time reconstruction of hand-object
  interactions.
\newblock {\em {Transactions on Graphics (TOG)}}, 40(3):29:1--29:12, 2021.

\bibitem{Zhang2019endtoend}
Xiong Zhang, Qiang Li, Hong Mo, Wenbo Zhang, and Wen Zheng.
\newblock End-to-end hand mesh recovery from a monocular {RGB} image.
\newblock In {\em {International Conference on Computer Vision ({ICCV})}},
  pages 2354--2364, 2019.

\bibitem{zhou2022toch}
Keyang Zhou, Bharat~Lal Bhatnagar, Jan~Eric Lenssen, and Gerard Pons-Moll.
\newblock {TOCH}: Spatio-temporal object-to-hand correspondence for motion
  refinement.
\newblock In {\em {European Conference on Computer Vision (ECCV)}}, volume
  13663, pages 1--19, 2022.

\bibitem{zhou2019rotation}
Yi Zhou, Connelly Barnes, Jingwan Lu, Jimei Yang, and Hao Li.
\newblock On the continuity of rotation representations in neural networks.
\newblock In {\em {Computer Vision and Pattern Recognition (CVPR)}}, pages
  5745--5753, 2019.

\bibitem{zhou2020monocular}
Yuxiao Zhou, Marc Habermann, Weipeng Xu, Ikhsanul Habibie, Christian Theobalt,
  and Feng Xu.
\newblock Monocular real-time hand shape and motion capture using multi-modal
  data.
\newblock In {\em {Computer Vision and Pattern Recognition (CVPR)}}, pages
  5345--5354, 2020.

\bibitem{ziani2022tempclr}
Andrea Ziani, Zicong Fan, Muhammed Kocabas, Sammy Christen, and Otmar Hilliges.
\newblock {TempCLR}: Reconstructing hands via time-coherent contrastive
  learning.
\newblock In {\em {International Conference on 3D Vision (3DV)}}, pages
  627--636, 2022.

\bibitem{zimmermann2017iccv}
Christian Zimmermann and Thomas Brox.
\newblock Learning to estimate {3D} hand pose from single {RGB} images.
\newblock In {\em {International Conference on Computer Vision ({ICCV})}},
  pages 4913--4921, 2017.

\bibitem{Freihand2019}
Christian Zimmermann, Duygu Ceylan, Jimei Yang, Bryan Russell, Max Argus, and
  Thomas Brox.
\newblock {FreiHAND}: {A} dataset for markerless capture of hand pose and shape
  from single {RGB} images.
\newblock In {\em {International Conference on Computer Vision ({ICCV})}},
  pages 813--822, 2019.

\end{thebibliography}
}

\newpage
{\large \textbf{Appendix - Supplementary Material}}
\vspace{2mm}
\setcounter{section}{0}
\setcounter{figure}{0}
\setcounter{table}{0}

\section{Dataset Details}
\label{sec:dataset}

\begin{figure}[b]
    \centering
    \includegraphics[width=0.8\linewidth]{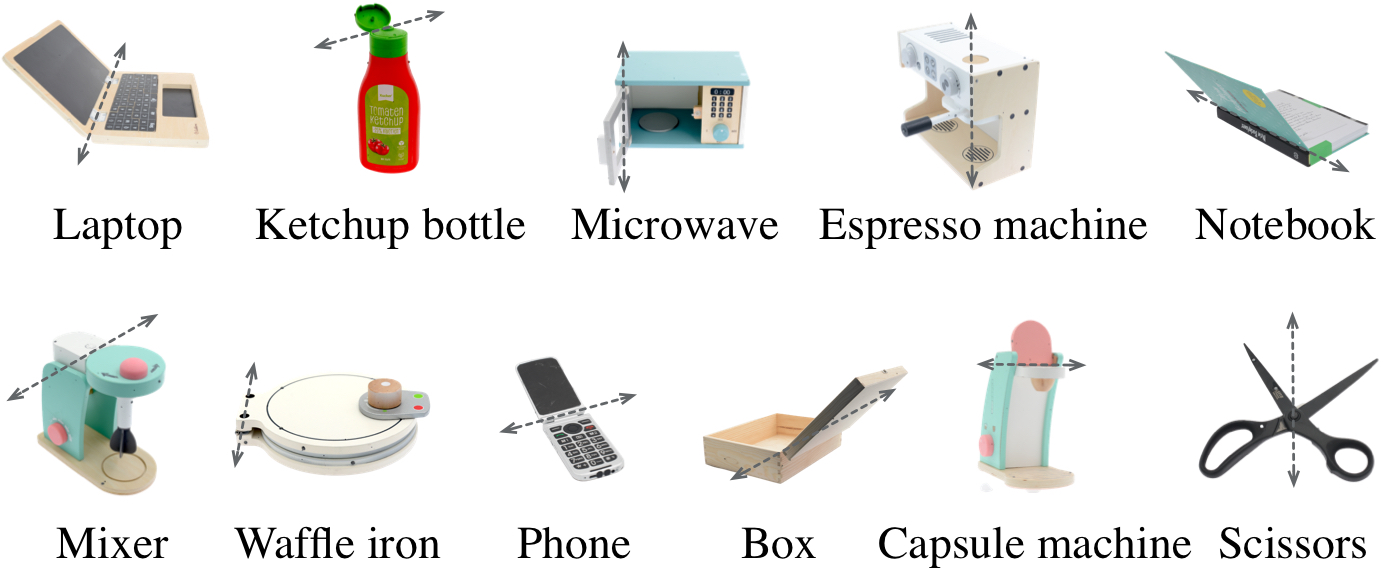}
    \caption{
     \subtitle{\datasetname objects} Each line shows the articulation axis.}
    \label{fig:objects}
\end{figure}

\myparagraph{Objects in \datasetname}
\refFig{fig:objects} shows all \numObjs articulated objects in our dataset.  Objects in \datasetname consists of two rigid parts that rotate about an axis. Each dash line  in the figure shows the articulation axis.

\begin{figure}[t]
    \centering
    \includegraphics[width=1.0\linewidth]{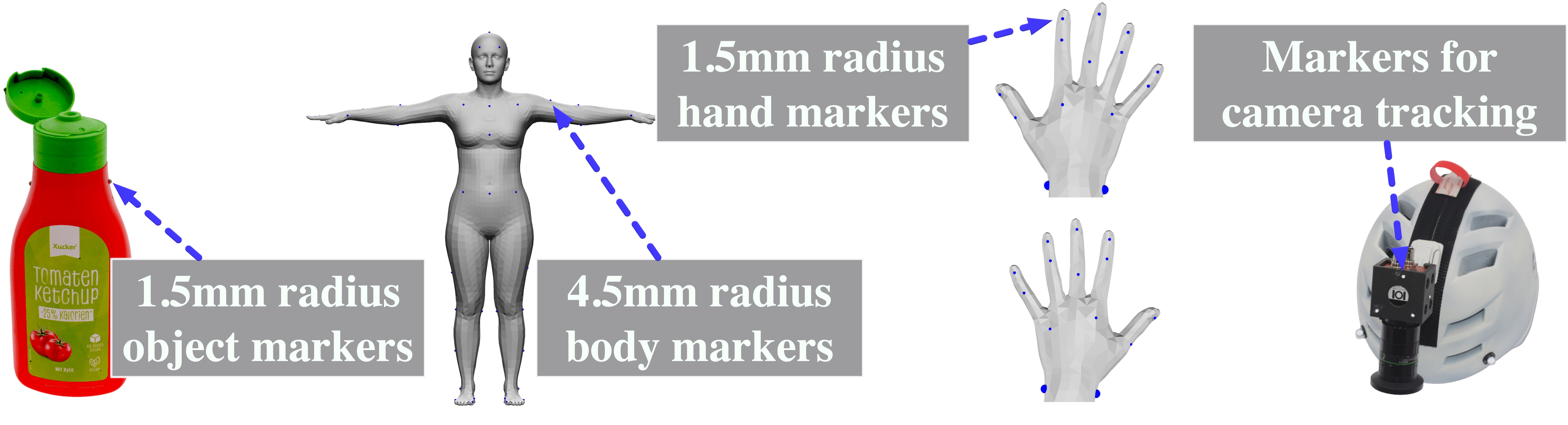}
    \caption{\subtitle{Markers for motion capture (\mocap)} We put \smallMarkersize radius markers on objects, hands and the egocentric camera. For the body, we use \mediumMarkersize radius markers. The markers are shown here to scale. Best viewed in color and zoomed-in.
    }
    \label{fig:capture_setup}
\end{figure}

\myparagraph{Marker sets}
\refFig{fig:capture_setup} shows marker sets for an object, the full human body, two hands, and the egocentric camera along with the marker size. Markers in this visualization are shown to scale.
The marker locations for all objects and all subjects can be found in the data release.

\myparagraph{Dataset statistics} 
\refTab{tab:subjects} shows the number of images and the number of sequences per subject for our dataset.
The average sequence length is $698$ frames (view-agnostic), corresponding to $23.3$ seconds.
In total, we have 2.1M images and there are more than 200k images for most subjects.
\refTab{tab:objects} shows the number of images per object. All objects have more than 170k images. 
To encourage different modes of interaction, we capture different intents for each object: ``use" and ``grasp".
Although both are for dexterous manipulation, in the ``use" sequences, the subjects are allowed to articulate the object but not in the ``grasp". 
Since we focus on studying  articulation, we capture more ``use" sequences.

\myparagraph{Protocol splits}
\refTab{tab:data_splits} shows the number of images and subjects in the allocentric and the egocentric settings.
Both settings use the same subject split -- 8 subjects for training, 1 for validation and 1 for testing.
The allocentric setting uses images from the 8 allocentric static views for training, validation, and testing.
The egocentric setting, in the training split, we allow models to use images from all $9$ views for additional supervision; 
During inference, however, models are evaluated with only egocentric images.

\begin{figure}[]
    \centering
    \includegraphics[width=1.0\linewidth]{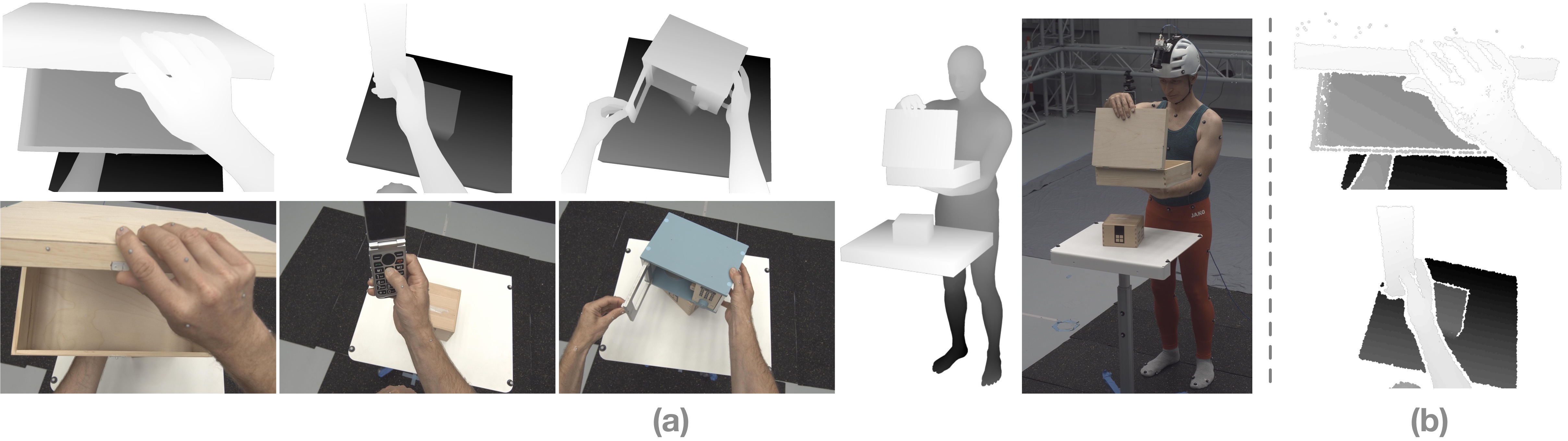}
    \caption{
    \textbf{Rendered depth images of human-object interaction.}
    (a) Rendered depth images of our 3D data, and the corresponding RGB images, and (b) depth images with synthetic kinect noise. Best viewed zoomed-in. For better depth-map visualization we do not render the floor here.}
    \label{fig:depth_images}
\end{figure}

\myparagraph{Depth images}
Since we perform full-body capture, we can render depth images with full-body interaction. 
Since most existing articulated object datasets contain neither two-hands nor human bodies\ccite{liu2022hoi4d,xiang2020sapien,michel2015pose}, and having a human in the scene is a realistic setting, we believe that \datasetname brings additional challenges of heavy occlusion and dynamic manipulation to the depth community.
\refFig{fig:depth_images} shows examples of the depth images and the corresponding RGB images.
The depth can be rendered with any synthetic sensor noise model (e.g.,~Kinect, right).

\begin{table}[b]
\centering
\resizebox{1.00\linewidth}{!}{
\begin{tabular}{cccccccccccc}
\toprule
Subjects  & S1      & S2      & S3      & S4      & S5      & S6     & S7     & S8     & S9      & S10 & Total     \\
\hline
\# Images & 209k & 224k & 220k & 212k & 227k & 228k & 191k & 280k & 208k & 135k & 2.1M\\
\# Seqs & 34 & 37 & 38 & 31 & 34 & 36 & 29 & 42 & 37 & 21 & 339 \\
\bottomrule
\label{tab:split_subjects}
\end{tabular}
}
\caption{\subtitle{Number of images and sequences for each subject}
The average sequence length is $698$ frames (view-agnostic), corresponding to $23.3$ seconds.
}
\label{tab:subjects}
\end{table}

\begin{table*}[t]
\centering
\resizebox{1.00\linewidth}{!}{
\begin{tabular}{ccccccccccccc}
\toprule
Objects & Notebook  & Box       & Espresso machine & Waffle iron & Laptop    & Phone     & Capsule machine & Mixer     & Ketchup bottle & Scissors  & Microwave & Total      \\
\hline
Use     & 163k & 152k  & 141k         & 166k    & 171k  & 159k  & 159k        & 156k & 138k       & 128k  & 144k  & 1.7M \\
Grasp   & 27k    & 36k   & 46k           & 41k      & 40k    & 49k   & 37k          & 47k    & 51k        & 44k   & 43k   & 0.4M  \\
\hline
Total   & 190k & 187k & 187k        & 207k   & 211k & 208k & 196k       & 202k & 189k      & 172k & 186k & 2.1M  \\
\bottomrule
\end{tabular}
}
\caption{\subtitle{Number of images for each object in \datasetname}
In \datasetname, we focus on studying hand interaction with object articulation.
Therefore, we capture more ``use" sequences in which subjects can articulate the object.
To encourage different modes of interaction, we also ask subjects to ``grasp" the objects without articulating the objects.
Since we focus on object articulation, we capture more data for ``use" than in ``grasp".
\label{tab:objects}
}
\end{table*}
\begin{table}[t]
\centering
\resizebox{0.80\linewidth}{!}{
\begin{tabular}{cccc}

\toprule
Splits & \# Train Images & \# Val Images  & \# Test Images  \\
\hline
allo     & 1.5M            & 202k          & 195k             \\
ego     & 1.7M         & 25k          & 24k \\
\bottomrule
\end{tabular}
}
\caption{\subtitle{Number of images for each protocol}
}
\label{tab:data_splits}
\end{table}

\section{Data Capture Details}
\label{sec:capture_details}
\begin{figure*}[t]
    \centerline{\includegraphics[width=1.0\linewidth]{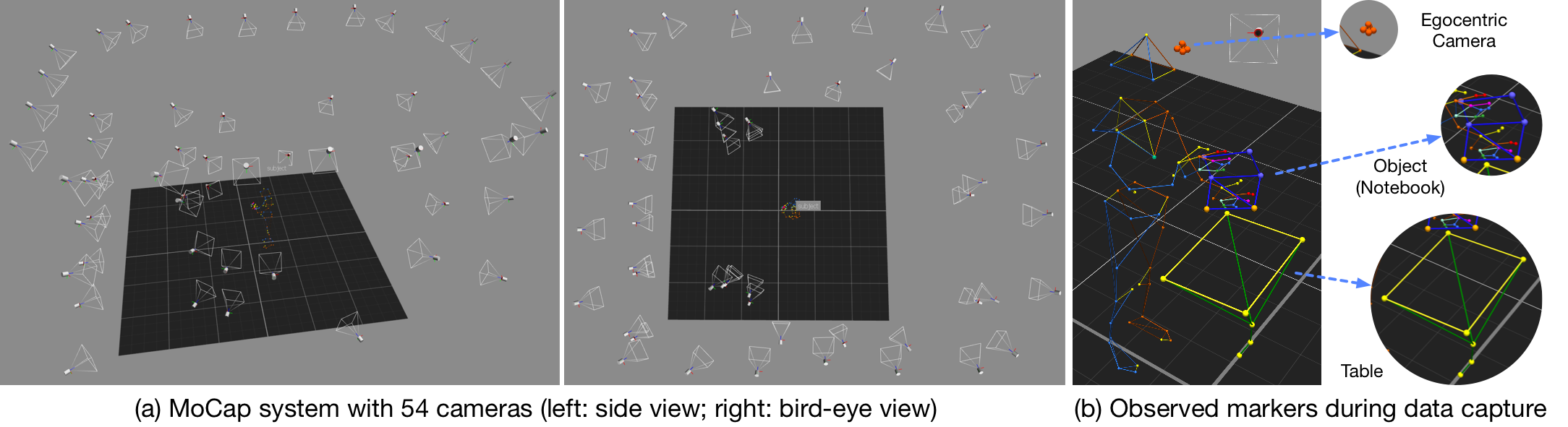}}
    \caption{
    \subtitle{Our capture system with \numVicon high-resolution \mocap cameras}
        (a) \mocap system in a side view and a bird-eye view, illustrating the \numVicon \mocap cameras used to eliminate occlusion during the capture.
        (b) Observed markers for a captured frame, showing markers tracking the full human body with hands, the object (notebook in this case), the egocentric camera, and props such as the table.
        Best viewed in color and zoomed in. 
    }
    \label{fig:vicon_system}
\end{figure*}%

\subsection{\mocap System with 54 Cameras}
\refFig{fig:vicon_system} illustrates our \mocap system. 
When capturing quality hand-object interaction data, the key is to eliminate occlusion in hand self-occlusion, hand-hand occlusion, and hand-object occlusion settings.
To minimize these sources of occlusion, we use \numVicon high-resolution \mocap cameras during our capture.
The camera positions and orientations are shown in \reffig{fig:vicon_system}a in a side view and a bird-eye view.
We also show the markers tracked by our system in \reffig{fig:vicon_system}b. 
The system tracks markers on the egocentric camera, the human subject (full body with hands), the articulated object (notebook in this case), and props such as the table.

\subsection{Creating Personalized Template} 
To create a personalized full body-and-hand template mesh for each subject, we obtain \threeD scans of the subjects in varying poses using a 3dMD scanner\ccite{3dmdhand}. 
We then register the \smplX model to the scans to obtain aligned meshes. 
The registered \smplX meshes are unposed to a canonical \tpose.
We perform a \smplX model-based fitting to the unposed meshes using vertex-to-vertex distances with the \smplX vertex correspondences.
Fitting with multiple {\tpose}d meshes allows us to filter out potential noise, and to capture the occluded regions of the body and hands, resulting in a reliable personalized \smplX template for each subject.

\subsection{Estimating Rotation Axis and Articulated Pose}
To solve for the rotation axis of each articulated object, we attach markers on each rigid part of each object and capture a calibration \mocap sequence by articulating its two parts.
Since the two parts rotate about an axis, the trajectory of each marker follows a circle on a \twoD plane.
We solve for the center of each circle using least-squares, and fit a \threeD line through the centers to obtain an initial rotation axis estimate.
We then refine the rotation axis estimate by minimizing a cost function.

Formally, let $X_t^i\in \R^3$ denote the \threeD position of a marker $i$ at time step $t$ placed on the ``top" part of the object (\eg, the lid of the ketchup bottle); $Y_t^j \in \R^3$ denotes a marker $j$ on the ``bottom" part of the object (\eg, the main body of the ketchup bottle) at time $t$. 
We pick a frame $t_0$ corresponding to a pre-defined rest pose for each object (for example, a frame in which the lid of the ketchup bottle is closed).
We then transform the \mocap sequence $\{X_t^i\}_{i, t}$ and $\{Y_t^j\}_{j, t}$ into a canonical sequence $\{\bar{X}_t^i\}_{i, t}$ and $\{\bar{Y}_t^j\}_{j, t}$ via a rigid transformation $(R, T)$ such that $\bar{Y}_t^j = Y_{t_0}^j$ for all $t$ and $j$.
After the canonicalization, the markers on the bottom part of the object are stationary across the entire sequence, and the top markers rotate around an axis.
Further, the trajectory of each marker is a circle on a \twoD plane.
We fit a circle to the trajectory $\{\bar{X}^{i}_t\}_{t=1, \cdots, N}$
 of an arbitrary marker $i$ using least-squares, and convert the \twoD circle center to  the \threeD space $(x_i, y_i, z_i)$.
We then fit a \threeD line $(v, v_0)$ to the center of each circle $\{(x_i, y_i, z_i)\}_i$ in \threeD, where $v\in\R^3$ is a unit directional vector for the line and $v_0\in \R^3$ is an arbitrary anchor point that the line crosses. 
Since fitting a \threeD line to \threeD centers can be imprecise, we refine the rotation axis further by minimizing the following cost function
\begin{equation}
\resizebox{1.00\linewidth}{!}{
$
    (v^*, v_0^*, \omega^*_{t=1,\cdots, N}) = \argmin_{\rule{-9ex}{3ex}v, v_0, \omega_{t=1,\cdots, N}} \sum_i ||\bar{X}_t^i - f(\bar{X}_{t_0}^i|v, v_0, \omega_t)||^2_2
$
}
\end{equation}
where $N$ is the number of frames in a sequence; $\omega^*_{t=1, \cdots, N}$ are the articulation angles in radians for all the frames relative to the rest pose.
The quantities $v^*, v_0^*$ are the refined rotation axis for the object.
The function $f(\bar{X}_{t_0}^i|v, v_0, \omega_t)$ rotates $\{\bar{X}^{i}_t\}$ about the estimated axis $(v^*, v_0^*)$ by $\omega_t$, the amount of articulation.
This ensures that the estimated rotation axis is consistent with the marker trajectory in the \mocap data.

To define the articulated object pose, we need to estimate the \oneD articulation angle, and its \sixD rigid pose.
To compute the former, after the rotation axis $(v^*, v_0^*)$ is estimated, for an actual \mocap sequence, the articulation angles are obtained by performing a 2D projection of the \threeD marker positions along the rotation axis.
Since the \twoD projection lies on a circle, the articulation angle can be estimated arithmetically.
The articulation angle is measured relative to each object's rest pose defined in $t_0$ during the rotation axis estimation step.
We take the median of the articulation angles estimated from all markers at a time step as our \groundtruth articulation angle.
Finally, to define the articulated object pose, we also need the \sixD object pose for its orientation and translation.
To solve for the \sixD pose $(R_t, T_t)$ for a frame $t$, we compute the rigid transformation from $\bar{Y}_{t_0}$ to $Y_t$. 
In other words, we compute the \sixD pose using the base marker according to its correspondence from the canonical space to the \mocap space.

\subsection{Computing Hand-Object Binary Contact}
We consider the two hands and the two parts of each articulated object as four watertight meshes for computing \groundtruth binary contact labels.
Given one mesh from the hands and one mesh from the object parts, we follow \grab\ccite{grab} to compute vertex-level contact. 
The main idea in \grab is to label vertices on a mesh as in contact with another mesh based on two cases: ``contact under-shooting" and ``contact over-shooting".
When vertices on a mesh are not inside another mesh, it is considered ``under-shooting", and geometric proximity is used to label contact.
When there is interpenetration between two meshes, for example, the thumb goes through a thin structure, the vertices of the thumb that ``over-shoot" the thin structure are labeled as in contact as well as the vertices that are inside the structure.
For more details, we refer readers to \ccite{grab}.

\section{Model Details and Results}
\label{ssec:training_model}
\label{sec:details_imp_arcticnet}
\myparagraph{General implementation details}
For all experiments, we use a ResNet-50\ccite{sun2019deep} backbone pre-trained on ImageNet\ccite{deng2009imagenet}.
The models are trained with the Adam optimizer\ccite{kingma2014adam} using a learning rate of $1e^{-5}$. 
For visibility, we crop each image around a square region centered around the object and resize the image to $224\times 224$.
Data augmentation is applied to the input image: rotation ($\pm 30^\circ$), scaling ($\pm 25\%$), and color jittering ($\pm 40\%$).

\begin{table}[]
\centering
\resizebox{1.00\linewidth}{!}{
\begin{tabular}{ccl}
\toprule
\multicolumn{3}{c}{\textbf{Hand Branch}}                                                     \\\hline
\textbf{Nr.} & \textbf{Module}          & \textbf{Details}                                   \\\hline
1            & pool                     & AvgPool2d(output\_size=1)                          \\
2            & cam\_init                & Linear(in\_dim=2048, out\_dim=512, bias=True)      \\
3            & cam\_init                & ReLU()                                             \\
4            & cam\_init                & Linear(in\_dim=512, out\_dim=512, bias=True)       \\
5            & cam\_init                & ReLU()                                             \\
6            & cam\_init                & Linear(in\_dim=512, out\_dim=3, bias=True)         \\
7            & refine.fwd               & Concat({[}``feat", ``hand\_pose", ``cam", ``shape"{]}) \\
8            & refine.fwd               & Linear(in\_dim=2157, out\_dim=1024, bias=True)     \\
9            & refine.fwd               & ReLU()                                             \\
10           & refine.fwd               & Dropout(p=0.5)                                     \\
11           & refine.fwd               & Linear(in\_dim=1024, out\_dim=1024, bias=True)     \\
12           & refine.fwd               & ReLU()                                             \\
13           & refine.fwd               & Dropout(p=0.5)                                     \\
14           & refine.decode.pose\_6d   & Linear(in\_dim=1024, out\_dim=96, bias=True)       \\
15           & refine.decode.shape      & Linear(in\_dim=1024, out\_dim=10, bias=True)       \\
16           & refine.decode.cam        & Linear(in\_dim=1024, out\_dim=3, bias=True)        \\\midrule
\multicolumn{3}{c}{\textbf{Object Branch}}                                                   \\\hline
1            & pool                     & AvgPool2d(output\_size=1)                          \\
2            & cam\_init                & Linear(in\_dim=2048, out\_dim=512, bias=True)      \\
3            & cam\_init                & ReLU()                                             \\
4            & cam\_init                & Linear(in\_dim=512, out\_dim=512, bias=True)       \\
5            & cam\_init                & ReLU()                                             \\
6            & cam\_init                & Linear(in\_dim=512, out\_dim=3, bias=True)         \\
7            & refine.fwd               & Concat({[}``feat", ``rot", ``cam", ``arti"{]})         \\
8            & refine.fwd               & Linear(in\_dim=2055, out\_dim=1024, bias=True)     \\
9            & refine.fwd               & ReLU()                                             \\
10           & refine.fwd               & Dropout(p=0.5)                                     \\
11           & refine.fwd               & Linear(in\_dim=1024, out\_dim=1024, bias=True)     \\
12           & refine.fwd               & ReLU()                                             \\
13           & refine.fwd               & Dropout(p=0.5)                                     \\
14           & refine.decode.rot  & Linear(in\_dim=1024, out\_dim=3, bias=True)        \\
15           & refine.decode.cam  & Linear(in\_dim=1024, out\_dim=3, bias=True)        \\
16           & refine.decode.arti & Linear(in\_dim=1024, out\_dim=1, bias=True)       \\
\bottomrule
\end{tabular}
}
\caption{\textbf{Details of the decoder in \methodnameSF.}}
\label{tab:arch_arctic_sf}
\end{table}

\subsection{\methodname}
\myparagraph{\methodnameSF Architecture}
We show the details of \methodnameSF in \refTab{tab:arch_arctic_sf}. 
\methodnameSF uses an encoder-decoder architecture.
Given an input image, we use average pooling to obtain a single image feature vector with dimension $2048$ from the backbone.
The image feature vector is used to predict the initial camera parameters (in ``cam\_init").
Following\ccite{Kanazawa2018_hmr}, we use an iterative refinement scheme to predict parameters of the hands and the objects.
First, we initialize all parameters to zero except for the camera parameters, which we initialize with the prediction in ``cam\_init".
For the hand branch, we concatenate the image feature vector, the initial hand poses, the hand shape vector into a single vector, and the predicted camera parameters for refinement (Line 7-16).
Within the refinement step, we first predict a latent vector using an MLP (Line 7-13). 
The latent vector is then being decoded via different heads to the residuals for MANO joint angles, shape and camera parameters (Line 14-16). 
The decoded residuals are added to the current estimates of the parameters respectively and will be used as inputs for the next refinement step (Line 7-16 again).
We have two iterations for the refinement. 
The object branch has a similar refinement scheme, but instead it predicts the object rotation, camera parameters, and object articulation.
Following \ccite{Kocabas_PARE_2021,mkocabas2021spec}, we use the \sixD rotation representation in\ccite{zhou2019rotation} for \mano joint angles.
Following\ccite{Boukhayma2019,Kanazawa2018_hmr,Kocabas_PARE_2021,Sarandi20FG,Zhang2019endtoend}, for the predicted weak perspective camera parameters, we use a fixed focal length of $1000.0$ and convert them to translations for each entity in the scene.

\myparagraph{\methodnameLSTM Architecture}
The LSTM model takes in a moving window of images and estimates \threeD meshes for each frame.
We use the same structure as \methodnameSF except that the image features within each window are passed through an LSTM network before being decoded.
We use a bidirectional LSTM with two hidden layers, hidden dimension of 1024, and a window size of $11$ based on validation.

\myparagraph{Training losses}
For each frame, our loss $\mathcal{L}$ is defined as the sum of the left hand,  right hand,  object, and interaction losses: $\mathcal{L} = \mathcal{L}_l +  \mathcal{L}_r +  \mathcal{L}_o + \mathcal{L}_{int}$.
In particular, the hand losses are defined as
\begin{equation}
    \mathcal{L}_{h} = \lambda_{3D}^{h} \mathcal{L}_{3D}^{h}+\lambda_{2D}^{h} \mathcal{L}_{2D}^{h}+\lambda_{\Theta}^{h}\mathcal{L}_{\Theta}^{h}
    +\lambda_{T}^{h} \mathcal{L}_{T}^{h} ,
\end{equation}
where $h = \{ l, r \}$ denotes the handedness. 
We fully supervise the \threeD joints (after subtracting the roots), the \twoD re-projection of the predicted \threeD joints, the \mano pose and shape parameters and the weak-perspective camera parameters.
Similarly, we pre-define \threeD landmarks for objects using farthest point sampling\ccite{eldar1994fps,eldar1997fps} on the object mesh. 
Using these landmarks, we formulate the object losses as
\begin{equation}
\mathcal{L}_{o} = \lambda_{3D}^{o} \mathcal{L}_{3D}^{o}+\lambda_{2D}^{o} \mathcal{L}_{2D}^{o}+\lambda_{\omega} \mathcal{L}_{\omega}+\lambda_{R} \mathcal{L}_{R}+\lambda_{T}^{o} \mathcal{L}_{T}^{o} ,
\end{equation}
where $\mathcal{L}_{\omega}$, $\mathcal{L}_{R}$ and $ \mathcal{L}_{T}^o$ supervise the articulation angle in radians, the global orientation and the weak perspective camera parameters. 
For the interaction loss $\mathcal{L}_{int}$, we use the contact deviation (CDev) metric (see main paper) as a loss term to improve hand-object contact.
We apply this loss between the left-hand/object, and right-hand/object. The loss $\mathcal{L}_{int}$ is a sum of the two.
All losses above use the MSE criterion.
All $\lambda$ variables are hyper-parameters and are set empirically based on validation performance.
In particular, we set all $\lambda$s to $1.0$ except $\lambda_{3D}^{*}=5.0$, $\lambda_{2D}^{h}=5.0$, $\lambda_{\Theta}^{h}=10.0$, $\lambda_{\beta}^{h}=0.001$ where $*$ denote a hand or an object.

\myparagraph{Training details}
We train with a batch size of $64$.
For the allocentric setting, we train single-frame models for $20$ epochs.
Since training temporal model is computation intensive, following VIBE~\cite{kocabas2020vibe}, we dump image features of pre-trained single-frame models to disk then train the LSTM models directly on the image features for $10$ epochs.
For the egocentric setting, since a model has access to both allocentric and egocentric images during training, to speed up training, we finetune pre-trained allocentric models on egocentric training images (1 camera)  for $50$ epochs.

\myparagraph{Camera model}
Following previous work on body and hand surface reconstruction \ccite{Boukhayma2019,Kanazawa2018_hmr,Kocabas_PARE_2021,Sarandi20FG,Zhang2019endtoend,mkocabas2021spec}, to estimate the translation of hands and objects ($T_l$, $T_r$, $T_o$), we predict weak-perspective camera parameters $(s, t_x, t_y)$ for each entity in the scene. 
The camera parameters consist of the scale $s\in \R$ and translation $(t_x, t_y) \in \R^2$ in pixel space and the translation can be recovered from $(s, t_x, t_y)$~\cite{Kocabas_PARE_2021,mkocabas2021spec} via:
\begin{equation}
T = (t_x, t_y, \frac{2f}{ws}) \in \R^3  \text{.}
\end{equation}
The terms $w$ and $f$ are the patch size and the focal length.
We do this for each ($T_l$, $T_r$, $T_o$).

\begin{figure*}[t]
    \centerline{\includegraphics[width=1.0\linewidth]{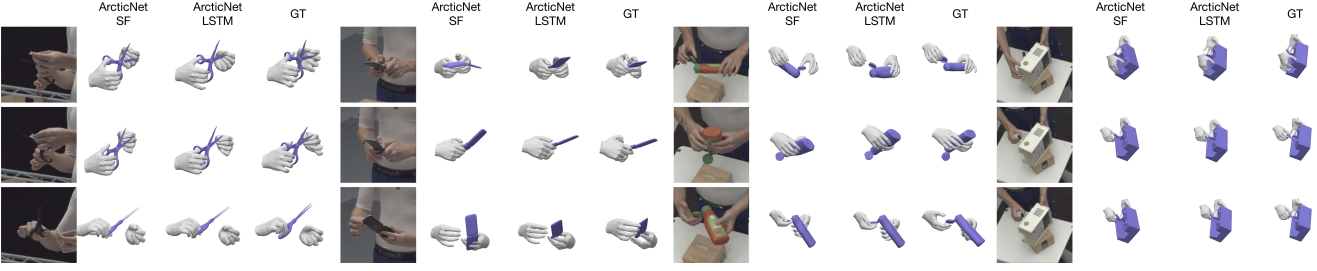}}
    \caption{
        \subtitle{Qualitative results of \methodnameSF and \methodnameLSTM} Best viewed in color and zoomed in. 
    }
    \label{fig:qualitative_results_sf_lstm}
\end{figure*}

\myparagraph{Qualitative Results}
\refFig{fig:qualitative_results_sf_lstm} shows the predictions of \methodnameSF and \methodnameLSTM on the test set.
As shown in the quantitative results in the main manuscript, the \methodnameLSTM model has lower errors overall for its prediction and it has better contact.
This is consistent with the observations in the qualitative examples here.
We hypothesize that this is because the LSTM allows the network to jointly reason between the motions of hands and objects.

\subsection{\intermethod}
\myparagraph{\interfieldSF Architecture}
\refTab{tab:arch_interfield} details our \intermethod model.
As an example, we illustrate how the right hand interaction field is predicted.
The left hand, and the object are predicted in a similar way.
In particular, from an input image, we obtain a $2048$-dimensional image feature vector from the image backbone.
The vector is passed through an MLP and is projected to lower dimension for computational efficiency (Line 1-4).
We use subsampled vertices of the hand, and concatenate the \threeD location of each vertex of the subsampled hand in the canonical pose with the $512$-dimensional image feature vector, resulting a point cloud with $515$ dimensions.
The point cloud is passed through a \pointnet backbone to obtain a latent point cloud with $512$ dimensions (Line 5-11).
Within the \pointnet backbone, the $515$-dimensional input point cloud is passed through a sequence of layers to produce lower level point features (Line 5-6).
The point features are further processed through Line 7-11.
We then concatenate the point cloud from the shallow layers (output of Line 6) and the deeper layers (output of Line 11) along the feature dimension, resulting in a point cloud whose individual points are in $1024$-dimensional.
A regressor maps each point ($1024$-dimensional) to a single scalar for distance prediction (Line 12-18).
Finally, we upsample the subsampled distances to the full hand mesh (Line 19).
We predict the interaction field of the left hand and the object in the same way. 
All entities shared the same image and \pointnet backbones. 
\begin{table}[]
\centering
\resizebox{1.00\linewidth}{!}{
\begin{tabular}{ccl}
\toprule
\textbf{Nr.} & \textbf{Module} & \textbf{Details}                     \\
\hline
1   & img\_feat.down  & Linear(in\_dim=2048, out\_dim=512, bias=True) \\
2   & img\_feat.down  & ReLU()                                        \\
3   & img\_feat.down  & Linear(in\_dim=512, out\_dim=512, bias=True)  \\
4   & img\_feat.down  & ReLU()                                        \\
5   & pointnet.shadow & Linear(in\_dim=515, out\_dim=512, bias=True)  \\
6   & pointnet.shadow & BatchNorm1d(512, affine=True)                 \\
7   & pointnet.deep   & Linear(in\_dim=515, out\_dim=512, bias=True)  \\
8   & pointnet.deep   & BatchNorm1d(512, affine=True)                 \\
9   & pointnet.deep   & ReLU()                                        \\
10  & pointnet.deep   & Linear(in\_dim=515, out\_dim=512, bias=True)  \\
11  & pointnet.deep   & BatchNorm1d(512, affine=True)                 \\
12  & regressor       & Linear(in\_dim=1024, out\_dim=512, bias=True) \\
13  & regressor       & BatchNorm1d(512, affine=True)                 \\
14  & regressor       & ReLU()                                        \\
15  & regressor       & Linear(in\_dim=512, out\_dim=128, bias=True)  \\
16  & regressor       & BatchNorm1d(128, affine=True)                 \\
17  & regressor       & ReLU()                                        \\
18  & regressor       & Linear(in\_dim=128, out\_dim=1, bias=True)    \\
19  & upsample      & Linear(in\_dim=195, out\_dim=778, bias=True)  \\
\bottomrule
\end{tabular}
}
\caption{\textbf{Details of \interfieldSF architecture.} }
\label{tab:arch_interfield}
\end{table}

\myparagraph{\interfieldLSTM Architecture}
The LSTM model takes in images from a window and estimates the interaction field for each frame.
In particular, we use the same architecture as in \methodnameSF except that we pass the image features in a window to an LSTM network before regressing the distances.
We use a bidirectional LSTM with two hidden layers, hidden dimension of $1024$ and a window size of $11$ based on validation performance.

\myparagraph{Training details}
For each frame, the network outputs are  $\hat{\boldsymbol{F}}{}^{l \rightarrow o}$, $\hat{\boldsymbol{F}}{}^{r \rightarrow o}$, $\hat{\boldsymbol{F}}{}^{o \rightarrow l}$, and $\hat{\boldsymbol{F}}{}^{o \rightarrow r}$.
To supervise training, we extract the ground-truth interaction fields for each frame from \datasetname and formulate an L1 loss 
$\mathcal{L} = \mathcal{L}_F(l, o) + \mathcal{L}_F(r, o) + \mathcal{L}_F(o, l) + \mathcal{L}_F(o, r)
$
where $\mathcal{L}_F(a, b) = || \boldsymbol{F}^{a \rightarrow b} - \hat{\boldsymbol{F}}{}^{a \rightarrow b} ||_1$ for entities $a$ and $b$.
For tractability and focus on close interaction, we threshold the interaction field distances at $10$cm for training and evaluation.

We train with a batch size of 64 for single-frame models and 32 for LSTM models. For the allocentric setting, we train single-frame models for $20$ epochs.
Since training temporal model has high computational requirements, following~\cite{kocabas2020vibe}, we dump image features of pre-trained single-frame networks to disk and train the LSTM models on the image features for $6$ epochs.
For the egocentric setting, a model has access to both allocentric and egocentric images in the training set. 
To speed up training, we finetune pre-trained allocentric models  on egocentric images (1 camera) for 100 epochs and 50 epochs for single-frame and LSTM models respectively.

\myparagraph{Qualitative Results}
\refFig{fig:qualitative_interfield_sf} shows the predictions of the single-frame model, and the corresponding \groundtruth. 
We use \groundtruth hand and object poses for visualization purposes. They are not the inputs of our network.
Here we focus on the colors on the meshes; Brighter colors represent closer distances in the interaction fields. 
The figure shows the feasibility of the task because the predictions correlate well with the \groundtruth.

\begin{figure}
    \centering
    \includegraphics[width=1.0\linewidth]{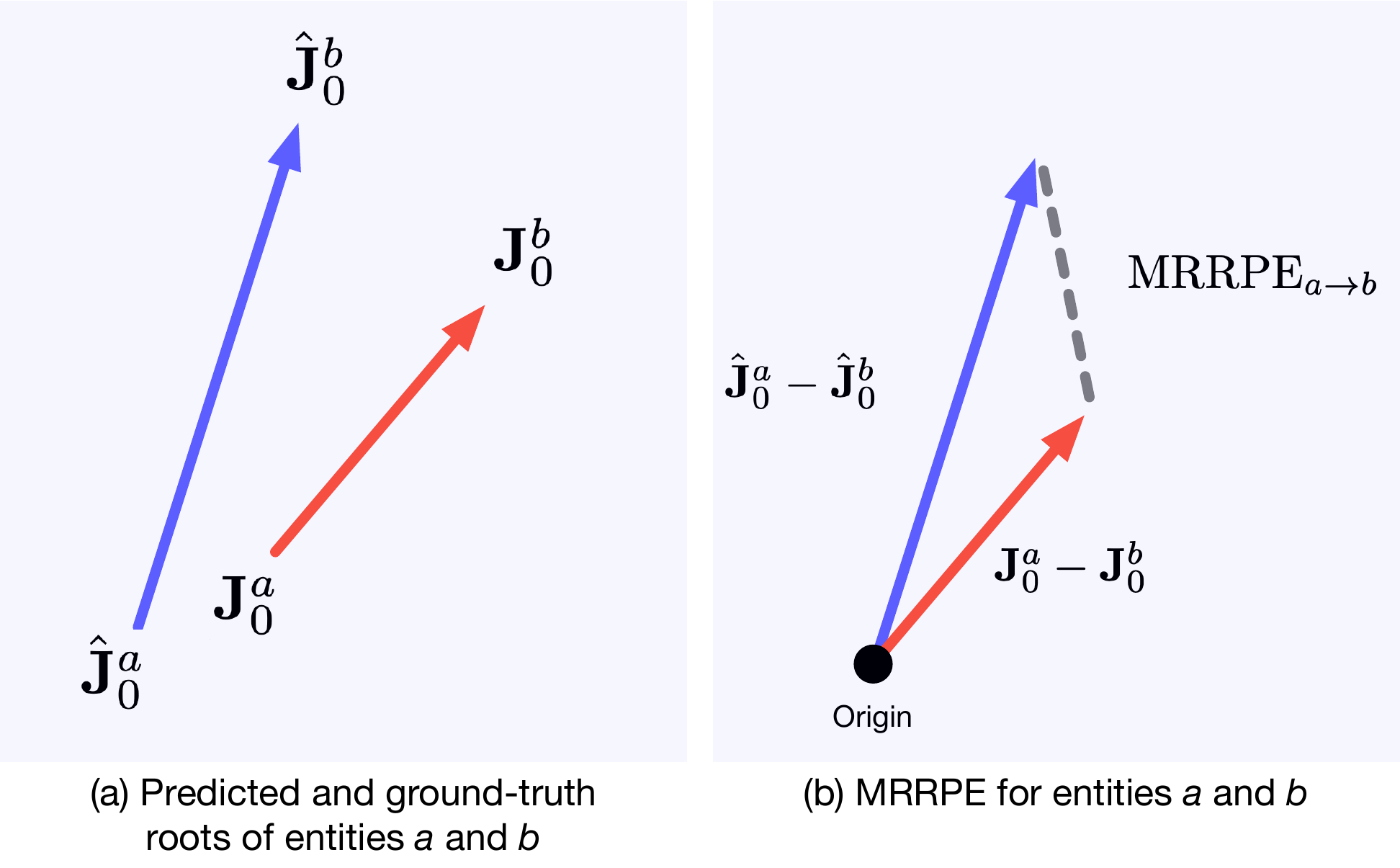}
    \caption{
     \subtitle{An illustration of the MRRPE$_{a\rightarrow b}$ metric}
     (a) The predicted roots of entities $a$ and $b$ are denoted by $\hat{\M{J}}_{0}^{a}$ and $\hat{\M{J}}_{0}^{b}$, and $\M{J}_{0}^{a}$ and $\M{J}_{0}^{b}$ are the corresponding \groundtruth. 
     (b) Subtract $a$ by $b$; MRRPE$_{a\rightarrow b}\in \R$ is indicated by the dash line.
     }
    \label{fig:mrrpe}
\end{figure}

\begin{table*}[t]
\resizebox{1.00\linewidth}{!}{
\begin{tabular}{cccccccc}
\toprule
                 & \multicolumn{2}{c}{Contact and Relative Position} & \multicolumn{2}{c}{Motion} & Hand    & \multicolumn{2}{c}{Object} \\
                 \hline 
Object           & CDev$_{ho}$ [$mm$] $\downarrow$ & MRRPE$_{rl/ro}$ [$mm$] $\downarrow$  & MDev$_{ho}$ [$mm$]  $\downarrow$ & ACC$_{h/o}$ [$m/{s^2}$] $\downarrow$ & MPJPE$_h$ [$mm$] $\downarrow$ & AAE [$^\circ$] $\downarrow$        & Success Rate [$\%$] $\uparrow$   \\\hline
Notebook         & 37.4                   & 47.7/39.8                & 9.9         & 5.0/6.5      & 20.8    & 3.3        & 80.4          \\
Box         & 47.5                   & \red{66.3}/\red{49.2}                & 10.6        & 5.5/6.7      & 24.5    & \blue{1.3}        & \blue{88.2}          \\

Espresso machine         & 48.9                   & 52.5/46.2                & 9.5         & 4.8/5.0      & 24.5    & \red{11.0}       & 81.0          \\
Waffle iron     & 41.8                   & 43.3/39.0                & \red{14.6}        & \red{5.6}/\red{7.9}      & 21.3    & 3.1        & 74.0          \\

Laptop          & 42.6                   & 54.7/40.5                & 12.8        & 5.2/7.2      & 21.7    & 1.7        & 84.4          \\
Phone           & 29.5                   & 42.2/31.1                & 7.5         & 4.6/7.2      & 18.8    & 3.9        & 62.3          \\
Capsule Machine & 30.5                   & \blue{37.6}/30.9                & 7.8         & 4.7/\blue{4.4}      & 19.2    & 6.9        & 69.3          \\
Mixer            & 34.5                   & 41.2/33.9                & 8.6         & 4.8/5.3      & 21.3    & 2.6        & 78.3          \\
Ketchup bottle   & 33.0                   & 45.6/35.0                & 10.8        & 5.4/7.4      & 20.7    & 7.0        & 59.2          \\
Scissors         & \blue{25.6}                   & 39.7/\blue{22.2}                & \blue{5.8}         & \blue{4.1}/5.0      & \blue{17.7}    & 10.5       & \red{50.1}          \\
Microwave         & \red{60.8}                   & 62.6/41.9                & 9.3         & 5.2/5.2      & \red{26.0}    & 7.3        & 74.3          \\
\bottomrule
\end{tabular}
}
\caption{\subtitle{Detailed breakdown on test set evaluation per object}
Here we provide the detailed breakdown of the test set evaluation according to each object.
For each metric, we use \red{red} to denote the object with the highest error; we use \blue{blue} to denote the lowest error.
\label{tab:sota_detail_p1_test}
}
\end{table*}
\section{Metrics and Experiments}
\label{sec:eval}
\subsection{Metric Details}
\myparagraph{Acceleration Error (ACC)}
Following\ccite{kocabas2020vibe}, we report acceleration error in $m/{s^2}$ to measure the smoothness of \taskpose, and \taskfield.
Formally, suppose $\hat{\V{h}}_i^t\in \R^d$ is the predicted vertex (or distance value) $i$ at frame $t$ of a hand; $\V{h}_i^t$ is the corresponding \groundtruth.
We compute the corresponding acceleration vector $\hat{\V{u}}_i^t\in \R^d$ of $\hat{\V{h}}_i^t$.
Similarly, we compute the acceleration vector $\V{u}_i^t$ for the \groundtruth.
The acceleration error for a hand is computed as:
\begin{equation}
\frac{1}{T V_h}\sum_{t=1}^{T} \sum_{i=1}^{V_h} \norm{\hat{\V{u}}_i^t - \V{u}_i^t}
\end{equation}
where $V_h$, $T$, $d$ are the number of hand vertices, the sequence length, and the number of dimension for the prediction of a task.
To compute the acceleration, we use centered difference:
\begin{equation}
    u_i^t = \frac{h_i^{t-1} -2 h_i^{t} + h_i^{t+1}}{w^2}
\end{equation}
where $w=1/30 s$ is the stencil width of 30-FPS videos.
Note that previous methods~\cite{kocabas2020vibe,ziani2022tempclr} computing the acceleration errors did not divide the error by $w^2$, leading to significantly smaller errors.
The prediction dimension ($d$) for the reconstruction task, and the interaction field task are $3$ and $1$ respectively.
For the former, we use root-relative vertices.
We compute the acceleration of the object in the same way.

\myparagraph{Average Articulation Error (AAE)} 
Suppose $\omega^t\in \R$ and $\hat{\omega}^t\in \R$ are the predicted and \groundtruth object articulation at frame $t$, the average articulation error is defined as
\begin{equation}
    \frac{1}{T}\sum_{t=1}^{T} |\omega^t - \hat{\omega}^t|
\end{equation}
where $T$ is the number of frames.

\myparagraph{Mean Relative-Root Position Error (MRRPE)} 
Following~\cite{interhand,fan2021digit}, to measure the relative root translation between two entities $a$ and $b$ in the scene (a hand or an object),
\begin{equation}
    \resizebox{0.7\hsize}{!}{%
        $\operatorname{MRRPE}_{a \rightarrow b}=\left\|\left(\M{J}_{0}^{a}-\M{J}_{0}^{b}\right)-\left(\hat{\M{J}}_{0}^{a}-\hat{\M{J}}_{0}^{b}\right)\right\|_{2} \text{,}$%
        }
\end{equation}
where $a\in \{l, r, o\}$ and $b\in \{l, r, o\}$ and $l, r, o$ denote the left hand, right hand, and the object, 
$\M{J}_{0}\in \R^3$ is the \groundtruth root joint location and $\hat{\M{J}}_{0}$ the predicted one.
\refFig{fig:mrrpe} shows a visualization of the metric.
Suppose we want to compute MRRPE$_{a\rightarrow b}$, and $\hat{\M{J}}_{0}^{a}$ and $\hat{\M{J}}_{0}^{b}$ denote the predicted roots for entities $a$ and $b$ and the notations without ``hat" are the \groundtruth.
The MRRPE value is the distance indicated in the dash line.

\subsection{Ablation and Analysis}
\begin{table}[]
\resizebox{1.00\linewidth}{!}{
\begin{tabular}{cccccccc}
\toprule
            & \multicolumn{2}{c}{Contact and Relative Position} & \multicolumn{2}{c}{Motion} & Hand    & \multicolumn{2}{c}{Object} \\\hline
Size & CDev$_{ho}$  &  MRRPE$_{rl/ro}$  &  MDev$_{ho}$     & ACC$_{h/o}$   & MPJPE$_h$ &  AAE & Success R. \\\hline
5           & 39.9                   & 48.0/37.4                & 9.3         & 6.2/8.2      & 23.3    & 6.1       & 72.7           \\
11          & \textbf{39.0}                   & \textbf{47.0/36.6}                & \textbf{8.8}         & \textbf{6.1}/7.7      & \textbf{22.8}    & \textbf{5.8}       & \textbf{74.6}           \\
15          & 39.7                   & 47.8/36.8                & 9.0         & 6.2/\textbf{7.6}      & 22.9    & \textbf{5.8}       & 74.4         \\
\bottomrule 
\end{tabular}
}
\caption{\subtitle{Effects of window size on \methodnameLSTM}
Here we ablate the effect of window size on our model on the validation set.
\label{tab:ablate_arcticnet_window}
}
\end{table}
\begin{table}[]
\resizebox{1.00\linewidth}{!}{
\begin{tabular}{cccccccc}
\toprule
            & \multicolumn{2}{c}{Contact and Relative Position} & \multicolumn{2}{c}{Motion} & Hand    & \multicolumn{2}{c}{Object} \\\hline
CDev loss & CDev$_{ho}$  &  MRRPE$_{rl/ro}$  &  MDev$_{ho}$     & ACC$_{h/o}$   & MPJPE$_h$ &  AAE & Success R. \\\hline
\xmark         & 49.0                   & 53.1/45.6                & 11.9        & 7.3/10.1      & \textbf{23.0}    & 6.1       & 71.3           \\
\cmark          & \textbf{41.9}                   & \textbf{50.1/37.6}                & \textbf{10.4}         & \textbf{7.3/9.8}      & 23.1    & \textbf{5.9}       & \textbf{71.8}           \\
\bottomrule 
\end{tabular}
}
\caption{\subtitle{Effects of contact deviation (CDev) as a training loss}
Here we ablate the effect of the contact deviation metric as a loss on \methodnameSF on the validation set.
\label{tab:ablate_cdev}
}
\end{table}
\myparagraph{Detailed analysis on test set}
To see a performance breakdown per object, \refTab{tab:sota_detail_p1_test} shows the evaluation of the \methodnameLSTM model evaluated on the test set of the allocentric split.
We use \red{red} to denote with the worst value for each metric and \blue{blue} to denote the best value.
We can see that the microwave is the hardest object in terms of hand reconstruction (see MPJPE) and contact (see CDev$_{ho}$).
This is because when opening the microwave door, the fingers are often heavily occluded.
In contrast, the scissors has the lowest hand reconstruction error because it is a small object.
In terms of estimating the articulation angle, the box, however, has the smallest error (see AAE). 
We hypothesize that the articulation angle is easy to observe because the box is the largest object.
In contrast, the espresso machine has a small handle, which can be occluded heavily, so it has the highest AAE error.
Finally, the scissors is the hardest object to reconstruct its pose (see success rate) because it is a small object and in some view its dark texture is similar to the background color.

\myparagraph{Window size for \methodnameLSTM}
\refTab{tab:ablate_arcticnet_window} shows the effect of window size on \methodnameLSTM in the validation set. 
With more frames in a window, overall the model has better performance.
To balance performance and efficiency, we use a window size of 11 for our model.

\myparagraph{Contact deviation (CDev) as training loss}
\refTab{tab:ablate_cdev} shows the effect of the contact deviation metric as a training loss to encourage hand-object contact using the allocentric protocol.
The results are evaluated on the validation set.
In this ablation, we train the \methodnameSF by turning on and off the contact deviation loss. The loss is applied between left-hand/object, and right-hand/object as described in \refsec{sec:details_imp_arcticnet}.
Results show that the CDev loss improves hand-object contact indicated by the CDev$_{ho}$ metric.

\myparagraph{Number of views for \methodnameSF}
Since \datasetname has only 1 egocentric view, we ablate the effect of the number of allocentric views.
When trained with 2, 4, 6, 8 views, randomly selected, MPJPEs for hands are $49.0mm$, $35.0mm$, $25.8mm$, $23.4mm$; the object success rates are $40.8\%$, $57.0\%$, $62.1\%$, $68.6\%$ on the validation set.

\begin{figure}[t]
    \centering
    \includegraphics[width=1\linewidth]{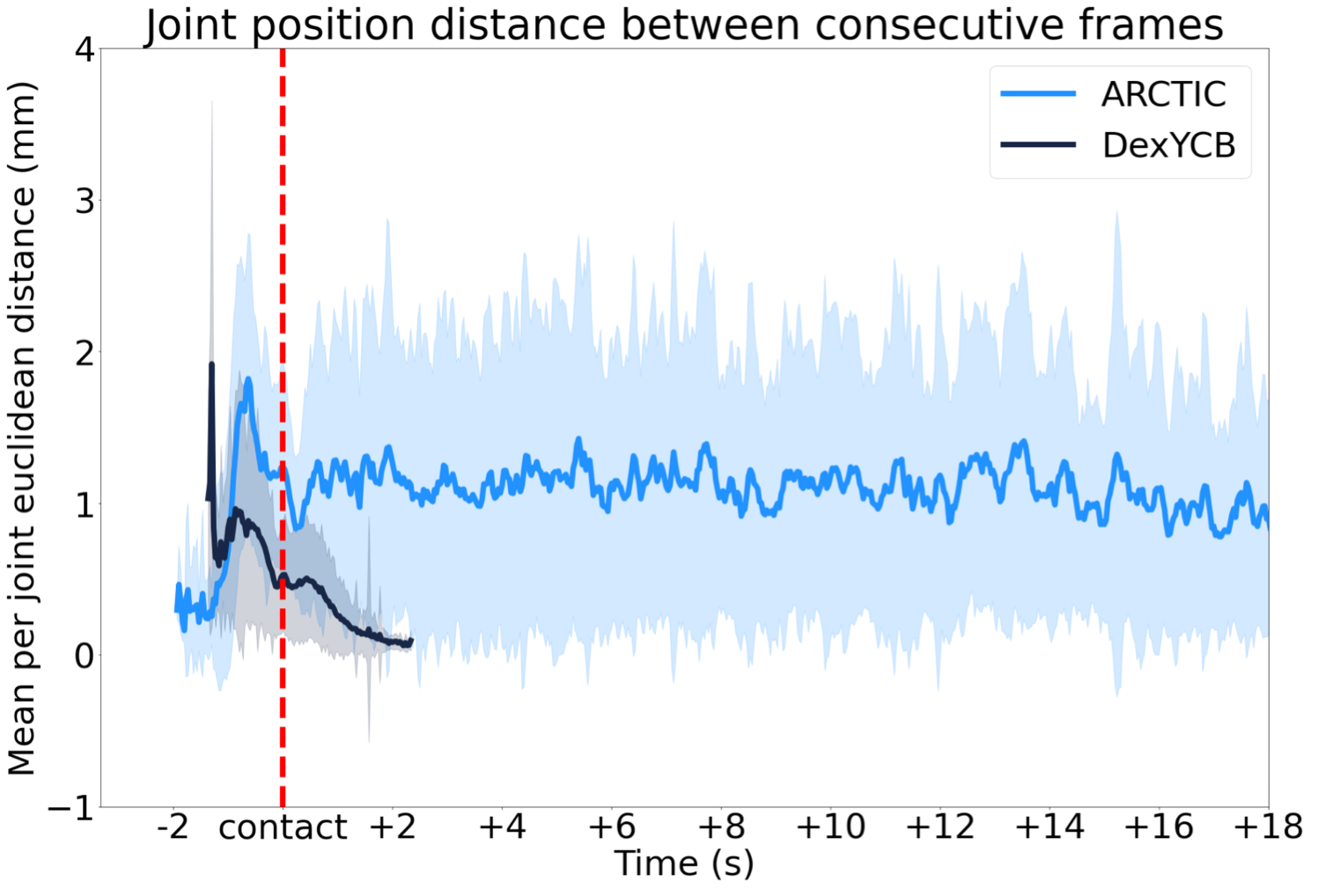}
    \caption{
    Changes in hand poses after first contact.
    }
    \label{fig:motion_plot}
\end{figure}

\myparagraph{Quantifying dexterous motion} 
Existing datasets do not show significant changes in hand pose.
In particular, poses in ContactPose are fixed relative to the object while DexYCB poses do not change much once contact is established (see \reffig{fig:motion_plot}). 
The figure plots the relative change in 3D joints across consecutive frames, without global translation and rotation.
The vertical dashed line indicates the first contact. 
Fig.~2 (main paper) shows that \datasetname has more diversity in hand poses and contact patterns, resulting from dexterous manipulation, compared to other datasets.

\begin{figure}[]
    \centering
    \includegraphics[width=1.0\linewidth]{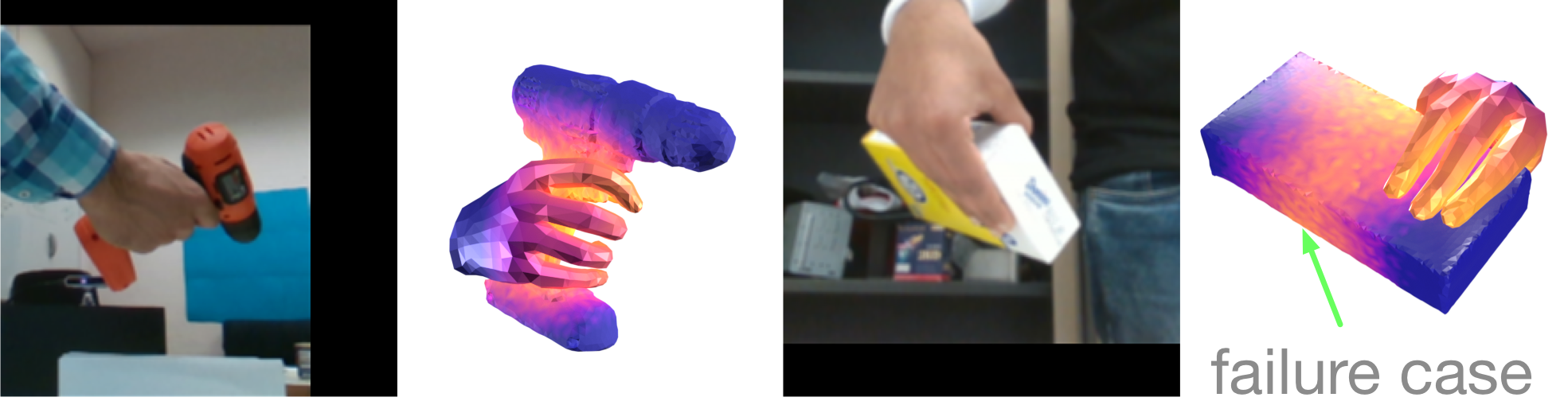}
    \caption{
    \textbf{Interaction field estimation on HO3D}}
    \label{fig:interfield_ho3d}
\end{figure}

\myparagraph{Interaction field estimation on HO3D} 
Our proposed task can be applied to existing hand datasets with rigid objects.
To show this, we trained InterField-SF on HO3D.
\reffig{fig:interfield_ho3d} shows qualitative results.
However, we note that existing hand-object datasets have similar hand poses within each sequence (thus, similar contact) and fewer training images, which are easy to overfit to (see \reffig{fig:interfield_ho3d} failure case); \datasetname is
large-scale and it is more challenging due to more dynamic
interaction (changing poses and contact) and thus will help
in fostering future research.

\myparagraph{Improve hand and rigid objects with ARCTIC} 
We pre-train \methodname on \datasetname, finetune on HO3D (hand + rigid objects).
This model is compared to a model trained only on HO3D.
Following the HO3D protocol, pre-training on \datasetname improves MPJPE (scale-translation aligned) errors by $9.2 \%$.
For the object, the vertex-to-vertex error (root aligned) improves by $7.1 \%$.
This shows that articulated hand-object data benefits hand and rigid object reconstruction.

\section{Visualizing \datasetname}

To visualize the our \threeD annotation in the dataset,
\reffig{fig:s_random_samples} shows random samples of the \threeD meshes of hands and objects overlaid on the images in our \datasetname dataset. 
See the \red{video} on our project page for our rendered sequences.

\section{Discussions and Limitations}
We introduce \datasetname, the first dataset for two hands dexterously manipulating articulated objects and baselines for the task of \taskpose, and \taskfield. Being a first step, our work is not without limitations.

\myparagraph{Known object models in \methodname}
Similar to existing methods~\cite{Yang_2021_CPF,hasson2019obman,Hasson2020photometric,Doosti2020hopeNet},
one limitation of our baselines is that they assume known object models. %
We view articulated \threeD shape estimation of unknown objects as an orthogonal problem on which the field is making progress. 
Now that we have showed the feasibility of inferring hand-object interaction for such objects, future work should bring together our method with \threeD articulated object inference. This is challenging %
and we believe it is critical to make progress on sub-problems, for which \datasetname can be leveraged.

\myparagraph{Toy objects}
Some of our objects are toys, %
which are not to scale and lack some of the visual complexity of real objects.
However, the aim of \datasetname is to study the physical dynamics between hand-object motions.

\myparagraph{SMPL-X for capturing contact}
Since we use SMPL-X/MANO as our human representation, the human geometry does not capture skin deformation during contact. 
While a deformable human body/hand model would be ideal for capturing true contact, developing such models is not a goal of \datasetname. 
Further, marker and image data \datasetname can be used to fit a deformable model if developed. 
Our 3D annotation and contact capture pipeline follows GRAB ~\cite{grab}.
In particular, we use MoSh++ to fit SMPL-X to the markers, producing highly accurate fits.
We adapt the contact capture pipeline from GRAB. 
GRAB contact labels are widely used in the community to support projects such as~\cite{zhou2022toch, taheri2021goal,grady2021contactopt}.

\myparagraph{Markers on hands in our \rgb images}
We use optical marker-based capture to provide accurate hand and object poses, thus potentially introducing label noise. %
However, our hand markers are minimally intrusive (\smallMarkersize in radius) and barely visible when  images are resized for inputs.

\myparagraph{Degree of freedom in our objects} 
We construct \datasetname with objects of 1 DoF.
This is because many items designed for human interaction often have a single axis of rotation as they are easy to produce and intuitive to manipulate (\eg, doors, refrigerators, ovens, pliers, \etc). Thus, objects in \datasetname are representative of a broad class of objects found in homes and businesses.
Importantly, reconstructing interaction with such objects involves occlusion, depth ambiguity, and contact estimation. 
These issues also apply to objects with more DoF.
Further, we capture more diverse hand poses and more challenging hand-object interactions compared to existing hand-object datasets.
Future work should expand the number and complexity of objects to further study the problems of depth ambiguity and occlusion.

\myparagraph{Approval for human subject data}
Subject data was collected with written, prior, informed consent and the data collection was reviewed by the university ethics board.

\begin{figure*}[]
    \centerline{\includegraphics[width=0.63\linewidth]{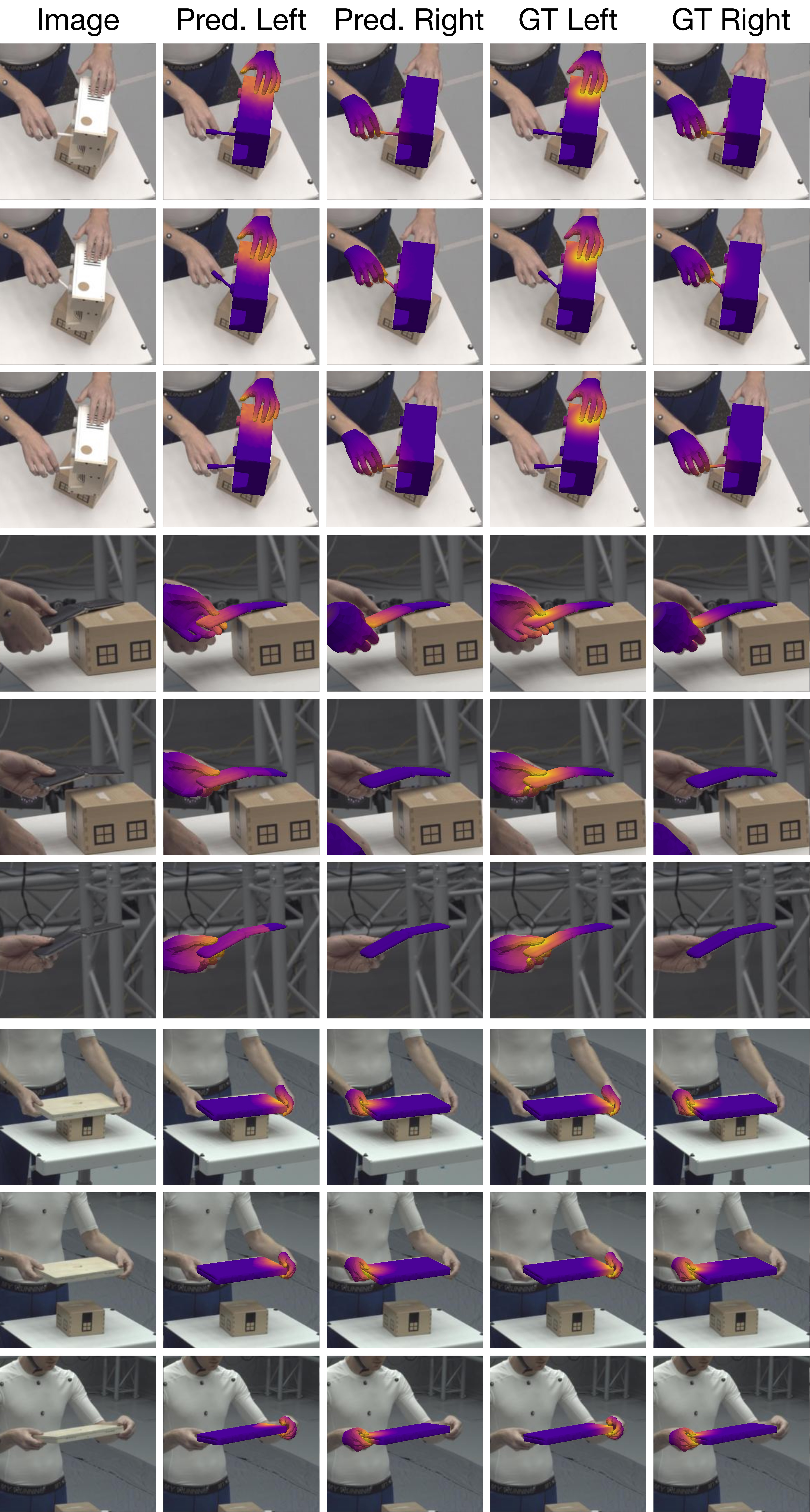}}
    \caption{
        \subtitle{Qualitative results of \interfieldSF} Best viewed in color and zoomed in. 
    }
    \label{fig:qualitative_interfield_sf}
\end{figure*}%

\begin{figure*}[t]
    \centering
    \vspace{2mm}
    \includegraphics[width=0.8\linewidth]{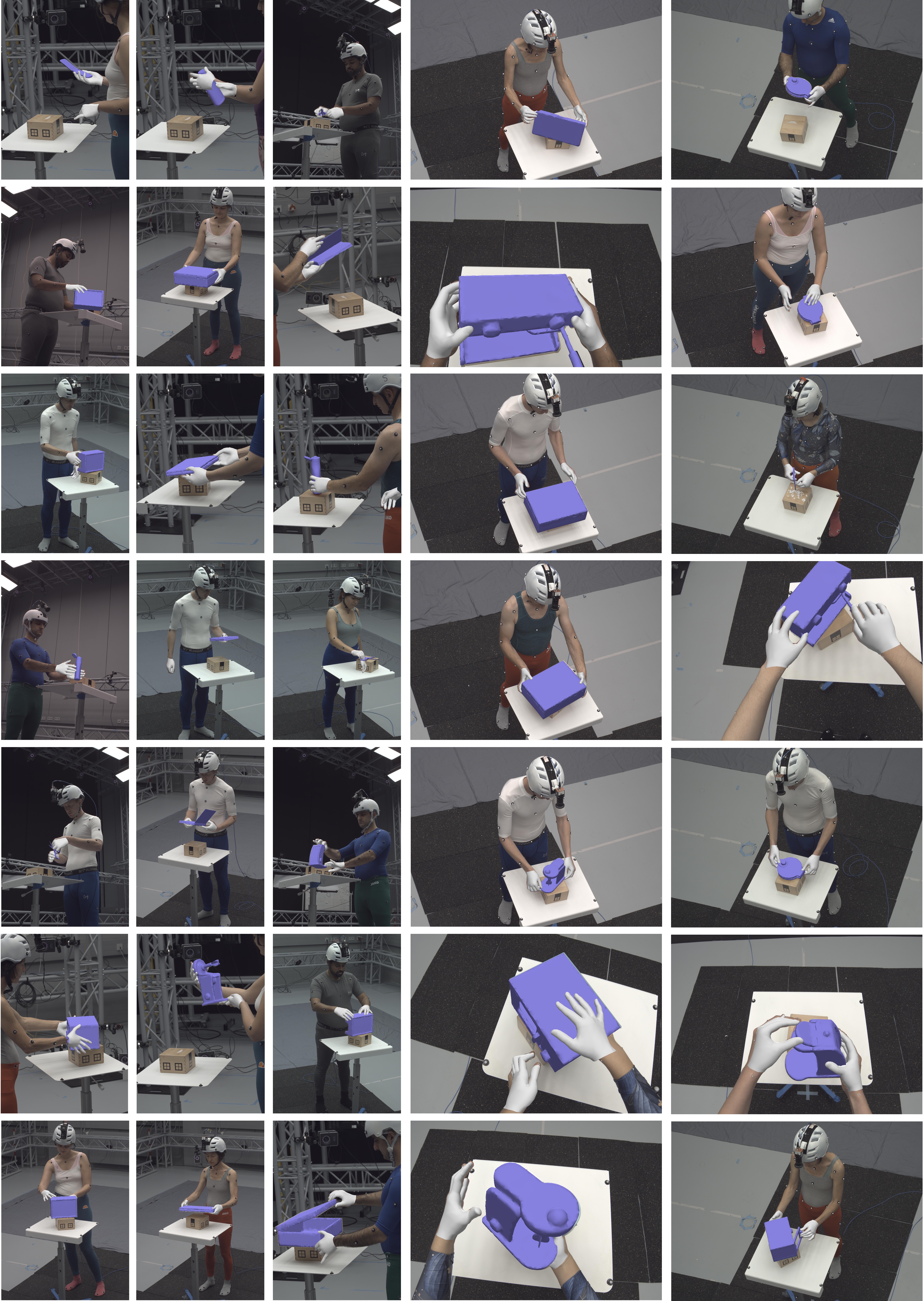}
    \caption{
    \textbf{Overlay of \datasetname\ \groundtruth} We overlay the \groundtruth in our dataset. Examples here are randomly sampled from \datasetname. Note that although the \rgb images contain both human bodies and hands, the hand region is clearly visible when zoomed in, thanks to our high resolution \rgb images.}
    \label{fig:s_random_samples}
\end{figure*}

\end{document}